\documentclass[pdflatex, onecolumn, sn-nature]{sn-jnl}
\usepackage{graphicx}%
\usepackage{multirow}%
\usepackage{amsmath,amssymb,amsfonts}%
\usepackage{amsthm}%
\usepackage[title]{appendix}%
\usepackage{xcolor}%
\usepackage{textcomp}%
\usepackage{manyfoot}%
\usepackage{booktabs}%
\usepackage{algorithm}%
\usepackage{algorithmicx}%
\usepackage{algpseudocode}%
\usepackage{listings}%

\usepackage{multirow, multicol, booktabs}
\usepackage{booktabs}
\usepackage{colortbl}

\usepackage{acro}

\usepackage[T1]{fontenc}
\usepackage[american]{babel} 
\usepackage{lmodern}
\usepackage{nicefrac}
\usepackage{cleveref}
\usepackage{nameref}
\usepackage{xfp}
\usepackage{siunitx}
\usepackage{mathtools} 

\usepackage{adjustbox}

\usepackage{soul}

\usepackage[protrusion=true,expansion=true]{microtype}
\usepackage{capt-of}  

\usepackage{fancyhdr}
\usepackage{enumitem}

\fancypagestyle{custompagestyle}{
  \fancyhf{}
  \fancyhead[LE,RO]{\small \leftmark} 
  \fancyhead[RE,LO]{\small Bertram et al.}        
  \fancyfoot[C]{\thepage}                        

}

\usepackage{chngcntr}
\newcommand{\giturl}{\url{https://github.com/bozeklab/actitect}}
\newcommand{\algoname}{{ActiTect}}
\newcommand{\algonamerbd}{{RBDisco}}

\newcommand{\pdrbd}{\textrm{PD}+\textrm{RBD}}
\newcommand{\pdnorbd}{\textrm{PD}–\textrm{RBD}}
\newcommand{\allrbd}{\textrm{all-RBD}}

\newcommand{\cgntrain}{\textrm{CogTrAiL-RBD}}
\newcommand{\cgntrainshort}{\textrm{CogTrAiL}}
\newcommand{\cgntest}{\textrm{Local Test}}
\newcommand{\oxflong}{\textrm{'Oxford Discovery' cohort from the Oxford Parkinson's Disease Centre (OPDC) project}}
\newcommand{\oxfshort}{\textrm{OPDC}}
\newcommand{\oxfirbd}{\textrm{OPDC (iRBD)}}
\newcommand{\oxfpdrbd}{\textrm{OPDC (\pdrbd)}}
\newcommand{\aar}{\textrm{PACE}}

\newif\ifphysstyle  
\physstyletrue    

\newif\ifshowstd
\showstdfalse      

\newcommand{\cidecimals}{3}  

\newcommand{\roundnum}[2]{\num[round-precision=#2,round-mode=places, retain-zero-exponent=false]{#1}}
\newcommand{\roundthreshold}{\fpeval{5 * 10^(-\cidecimals - 1)}}

\DeclareRobustCommand{\conditionalround}[1]{%
  \dimen0=\fpeval{abs(#1)}pt %
  \ifdim \dimen0 < \roundthreshold pt
    \roundnum{\fpeval{#1}}{\cidecimals+1}%
  \else
    \roundnum{\fpeval{#1}}{\cidecimals}%
  \fi
}

\newcommand{\ci}[4]{%
  \ifphysstyle
    $\conditionalround{#1}^{\scriptstyle \,+\conditionalround{\fpeval{#3 - #1}}}_{\scriptstyle \,-\conditionalround{\fpeval{#1 - #2}}}%
    \ifshowstd \newline(\sigma = \roundnum{#4}{\cidecimals})\fi$%
  \else
    $\conditionalround{#1}\,[\conditionalround{#2},\,\conditionalround{#3}]%
    \ifshowstd \,(\sigma = \roundnum{#4}{\cidecimals})\fi$%
  \fi
}

\newcommand{\withcidecimals}[2]{%
  \begingroup
    \def\cidecimals{#1}%
    \edef\roundthreshold{\fpeval{5 * 10^(-\cidecimals - 1)}}%
    #2
  \endgroup
}

\newcommand{\ciOne}[4]{\withcidecimals{1}{\ci{#1}{#2}{#3}{#4}}}
\newcommand{\ciTwo}[4]{\withcidecimals{2}{\ci{#1}{#2}{#3}{#4}}}
\newcommand{\ciThree}[4]{\withcidecimals{3}{\ci{#1}{#2}{#3}{#4}}}

\newcommand{\new}[1]{\textcolor{blue}{#1}}
\renewcommand{\new}[1]{#1} 

\usepackage{xparse}
\usepackage{xifthen}
\usepackage{xstring}
\usepackage{etoolbox}
\usepackage{nameref} 

\NewDocumentCommand{\optionalpar}{O{true} m}{%
  \ifstrequal{#1}{true}{#2}{}%
}

\makeatletter
\newcommand{\storeparname}[2]{%
  \protected@write\@auxout{}{%
    \string\gdef\string\parname@#1{#2}%
  }%
}
\makeatother
\newcommand{\getparname}[1]{\csname parname@#1\endcsname}

\newcommand{\parref}[1]{%
  \hyperref[#1]{Sec.~\ref{#1}}
  \ignorespaces        
}

\NewDocumentCommand{\newpar}{O{true} O{true} m O{}}{%
  \IfEqCase{#1}{{true}{\par\vspace{.5em}}}%
  \IfEqCase{#2}{{true}{\textbf{#3}}{false}{#3}}[\PackageError{newpar}{Invalid bold option '#2'}{}]%
  \ifstrempty{#4}{}{%
    \phantomsection%
    \label{#4}%
    \storeparname{#4}{#3}%
  }%
  \par\nopagebreak%
}

\definecolor{TableColor1}{gray}{0.9}
\definecolor{TableColor2}{rgb}{1, 1, 1}

\newcommand{\SuppItem}[2][]{%
  \item \hyperref[#2]{\textbf{#1:} \nameref{#2}\ (p.~\pageref{#2})}%
}

\makeatletter
\renewcommand{\abstract}[1]{%
  \def\@abstract{%
    \abstractfont%
    \textbf{\abstractname. }#1\par}%
}
\makeatother

\begin{document}

\title{\textbf{\algoname{}: A Generalizable Machine Learning Pipeline for REM Sleep Behavior Disorder Screening through Standardized Actigraphy}}

\author*[1,2,3]{\fnm{David} \sur{Bertram}}\email{dbertram@uni-koeln.de}
\author[4,5]{\fnm{Anja} \sur{Ophey}}
\author[5,6]{\fnm{Sinah} \sur{R\"ottgen}}
\author[7,8]{\fnm{Konstantin} \sur{Kufer}}
\author[6]{\fnm{Nele} \sur{Merten}}
\author[5,6]{\fnm{Gereon R.} \sur{Fink}}
\author[4]{\fnm{Elke} \sur{Kalbe}}
\author[10]{\fnm{Clint} \sur{Hansen}}
\author[10]{\fnm{Walter} \sur{Maetzler}}
\author[11,12]{\fnm{Maximilian} \sur{Kapsecker}}
\author[12]{\fnm{Lara M.} \sur{Reimer}}
\author[12]{\fnm{Stephan} \sur{Jonas}}
\author[13,15]{\fnm{Andreas T.} \sur{Damgaard}}
\author[13]{\fnm{Natasha B.} \sur{Bertelsen}}
\author[13]{\fnm{Casper} \sur{Skjaerbaek}}
\author[13,14]{\fnm{Per} \sur{Borghammer}}
\author[16]{\fnm{Karolien} \sur{Groenewald}}
\author[16]{\fnm{Pietro-Luca} \sur{Ratti}}
\author[16]{\fnm{Michele T.} \sur{Hu}}
\author[2,3]{\fnm{No\'{e}mie} \sur{Moreau}}
\author*[5,6,7,8]{\fnm{Michael} \sur{Sommerauer}}\email{michael.sommerauer@ukbonn.de}
\equalcont{Equal contribution.}
\author*[2,3,9]{and \fnm{Katarzyna} \sur{Bozek}}\email{k.bozek@uni-koeln.de}
\equalcont{Equal contribution.}

\affil[1]{\orgdiv{Faculty of Mathematics and Natural Sciences}, \orgname{University of Cologne}, \orgaddress{\country{Germany}}}
\affil[2]{\orgdiv{Institute for Biomedical Informatics}, \orgname{Faculty of Medicine and University Hospital Cologne, University of Cologne}, \orgaddress{\country{Germany}}}
\affil[3]{\orgdiv{Center for Molecular Medicine Cologne (CMMC)}, \orgname{Faculty of Medicine and University Hospital Cologne, University of Cologne}, \orgaddress{\country{Germany}}}
\affil[4]{\orgdiv{Medical Psychology\,|\,Neuropsychology and Gender Studies}, \orgname{Faculty of Medicine and University Hospital Cologne, University of Cologne},\orgaddress{ \country{Germany}}}
\affil[5]{\orgdiv{Cognitive Neuroscience, Insitute for Neuroscience and Medicine, INM-3}, \orgname{Research Center Juelich}, \orgaddress{\country{Germany}}}
\affil[6]{\orgdiv{Department of Neurology}, \orgname{Faculty of Medicine and University Hospital Cologne, University of Cologne}, \orgaddress{\country{Germany}}}
\affil[7]{\orgdiv{Center of Neurology, Department of Parkinson, Sleep and Movement Disorders}, \orgname{University Hospital Bonn, University of Bonn}, \orgaddress{\country{Germany}}}
\affil[8]{\orgdiv{German Center for Neurodegenerative Diseases (DZNE)}, \orgname{Bonn}, \orgaddress{\country{Germany}}}
\affil[9]{\orgdiv{Cluster of Excellence for Aging and Aging-Associated Diseases (CECAD)}, \orgname{University of Cologne}, \orgaddress{\country{Germany}}}
\affil[10]{\orgdiv{Department of Neurology}, \orgname{University Medical Center Schleswig-Holstein, Campus Kiel and Kiel University}, \orgaddress{\country{Germany}}}
\affil[11]{\orgdiv{Department of Informatics}, \orgname{Technical University of Munich}, \orgaddress{\country{Germany}}}
\affil[12]{\orgdiv{Institute for Digital Medicine}, \orgname{University Hospital Bonn}, \orgaddress{\country{Germany}}}
\affil[13]{\orgdiv{Lundbeck Foundation Parkinson's Disease Research Center (PACE)}, \orgname{Aarhus University}, \country{Denmark}}
\affil[14]{\orgdiv{Department of Nuclear Medicine}, \orgname{Aarhus University Hospital}, \country{Denmark}}
\affil[15]{\orgdiv{Department of Electrical and Computer Engineering}, \orgname{Aarhus University}, \country{Denmark}}
\affil[16]{\orgdiv{Oxford Parkinson's Disease Centre and Division of Neurology, Nuffield Department of Clinical Neurosciences}, \orgname{University of Oxford}, \country{UK}}

\abstract{
	Isolated rapid eye movement sleep behavior disorder (iRBD) is a major prodromal marker of $\alpha$‑synucleinopathies, often preceding the clinical onset of Parkinson’s disease, dementia with Lewy bodies, or multiple system atrophy.
	While wrist-worn actimeters hold significant potential for detecting RBD in large-scale screening efforts by capturing abnormal nocturnal movements, they require a reliable and efficient analysis pipeline.
	This study presents \algoname{}, a fully automated, open-source machine learning tool to identify RBD from actigraphy recordings.  
    To ensure generalizability across heterogeneous acquisition settings, our pipeline includes robust preprocessing and automated sleep–wake detection to harmonize multi-device data and extract physiologically interpretable motion features.
    Model development was conducted on a cohort of 78 individuals, yielding strong discrimination under nested cross-validation (AUROC = 0.95).
    Generalization was confirmed on a blinded local test set (n = 31, AUROC = 0.86) and two independent external cohorts (n = 113, AUROC = 0.84; n = 57, AUROC = 0.94).
    To assess robustness, leave-one-dataset-out cross-validation across cohorts demonstrated consistent performance (AUROC range = 0.84–0.89).
    Complementary stability analysis showed that predictive features remained reproducible across datasets, supporting the pooled multi-center pre-trained model for broader deployment.
    As an open-source, easy-to-use tool, \algoname{} promotes adoption, independent validation, and collaborative improvements, thereby advancing generalizable wearable-based RBD detection.
    }
    

\newgeometry{textwidth=45pc, top=1mm, bottom=20mm, centering}

\maketitle

\restoregeometry
\newpage 
\pagestyle{custompagestyle}

\section{Introduction}
Neurodegenerative disorders are a leading cause of illness and disability, affecting tens of millions of people worldwide with a significant increase in prevalence over the past three decades\,\cite{feigin_global_2020, dorsey_emerging_2018}.
Early detection, particularly during prodromal stages when clinical burden is low, is crucial for understanding disease onset mechanisms and enabling timely interventions\,\cite{postuma_prodromal_2019}.
A major subset of neurodegenerative disorders is driven by $\alpha$-synucleinopathies, which are characterized by abnormal accumulation of $\alpha$-synuclein in the nervous system and underlie the pathophysiology of conditions such as Parkinson’s disease (PD), dementia with Lewy bodies (DLB), and multiple system atrophy (MSA)\,\cite{spillantini_alpha-synuclein_1997, goedert_100_2013, brettschneider_progression_2017}.
These conditions share a common clincal marker of the prodromal phase: rapid eye movement (REM) sleep behavior disorder (RBD), a parasomnia characterized by loss of muscle atonia, abnormal movements and dream enactment behaviors during REM sleep\,\cite{berg_mds_2015, heinzel_update_2019, boeve_pathophysiology_2007, dauvilliers_rem_2018}.
In the absence of a clinical diagnosis of PD, DLB or MSA, this condition is termed isolated or idiopathic RBD (iRBD), which often represents an early manifestation of an underlying $\alpha$-synucleinopathy.
In fact, longitudinal studies have demonstrated that iRBD can precede the clinical onset of motor or cognitive symptoms in $\alpha$-synucleinopathies by up to $20\,\textrm{years}$, underlining its critical value as a predictive marker in both, clinical and research settings\,\cite{postuma_prodromal_2019, fereshtehnejad_evolution_2019}.

Currently, the gold standard for detecting RBD is video-polysomnography (vPSG), an accurate but resource-intensive diagnostic test, requiring costly equipment and expert manual analysis\,\cite{cesari_video-polysomnography_2022, gagnon_rapid-eye-movement_2006}, which in turn limits its practicality for large-scale application.
Clinical questionnaires on RBD symptoms provide a simpler and more practical alternative but suffer from subjectivity and reduced diagnostic accuracy\,\cite{sateia_international_2014, halsband_rem_2018, li_diagnostic_2017, stiasny-kolster_diagnostic_2015, chahine_questionnaire-based_2013}.\\
Actigraphy, typically utilizing wrist-worn accelerometers in RBD studies, has been proposed as a cost-effective, minimally intrusive, and quantitative approach for assessing RBD, making it particularly suitable for large-scale pre-screening.
Early studies demonstrated promising results but reported inconsistent performance\,\cite{louter_actigraphy_2014, naismith_relationship_2010, stefani_screening_2018}, likely due to limited number of features assessed and the subjectivity of human raters.\\
To overcome these limitations, machine learning (ML) approaches have been applied to actigraphy data\,\cite{brink-kjaer_ambulatory_2023, raschella_actigraphy_2023, brink-kjaerFullyAutomatedDetection2023}, enabling multivariate feature analysis and capturing complex patterns that univariate approaches may overlook.
While earlier investigations were largely conducted in single-center settings without independent test sets, more recent work has begun to address multi-center and multi-device validation.\,\cite{zhouActigraphybasedDetectionIsolated2025}\\
However, translating these advances into scalable clinical tools requires not only broader validation, but also transparent and reproducible implementations, unified end-to-end pipelines that operate across heterogeneous acquisition settings, and methodological frameworks that are robust to cohort and device variability while leveraging the richness of calibrated full-resolution tri-axial actigraphy data.
As multi-center efforts expand, progress will increasingly depend on moving beyond parallel model development toward integrative benchmarking and pooled learning across centers.
Openly available, easy-to-use implementations and harmonized methodological frameworks will be essential to enable such convergence, supporting community-driven systematic comparison and cumulative refinement, and ultimately broader reproducible deployment.

\begin{figure}[ht]
	\centering
	\includegraphics[page=1, width=\textwidth]{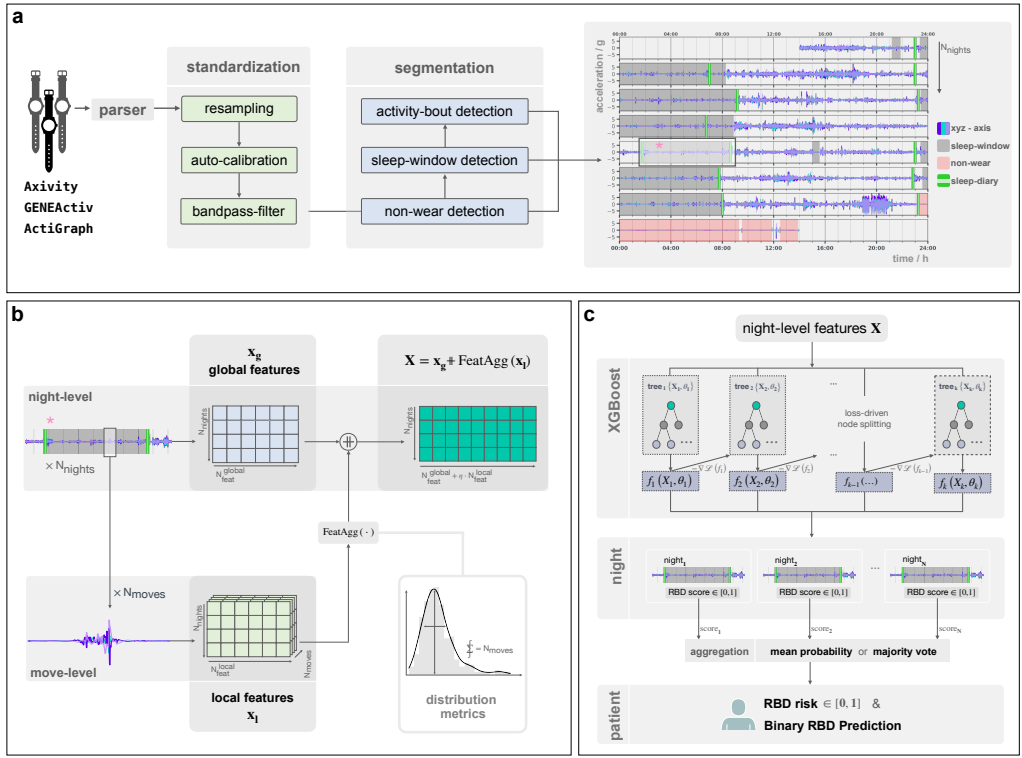}
	\caption{\label{fig:pipeline_overview}
    \textbf{\algoname{} pipeline overview.}
    \textbf{(a) Preprocessing.} 
    Raw actigraphy data from different devices is standardized through a dedicated preprocessing module, which mitigates systematic differences in signal distribution and enables generalizable motion feature extraction for downstream tasks. The pipeline further performs automated detection of sleep periods and non-wear episodes, reducing the need of manual annotations and enabling consistent analysis across large-scale datasets.
    \textbf{(b) Feature Extraction.} 
    From detected sleep bouts, we extract meaningful motion features that characterize nocturnal activity patterns relevant to RBD. Local features are computed for each activity bout, then aggregated to derive global descriptors representing the entire night.
    \textbf{(c) Predictive Model.}
    Each night’s extracted global motion features are mapped to an RBD probability score using boosted decision trees (XGBoost). These nightly scores are then aggregated into a patient-level risk score via a custom function that combines mean-probability thresholding and majority voting. The final binary RBD prediction is obtained by thresholding each patient’s aggregated risk score.}
\end{figure}

In this work, we present \algoname{} (\cref{fig:pipeline_overview}), an open-source ML tool for predicting RBD status from wrist-worn actigraphy, designed to facilitate cross-device harmonization, pretrained for immediate use, and optimized for computational efficiency and adaptability.
It consists of two components: a non-ML module for robust data handling and preprocessing, enabling data standardization across devices, and an ML-based model for classification.\\
The data processing module (\cref{fig:pipeline_overview}a) is compatible with common wrist-actigraphy devices and applies a series of operations to mitigate systematic artifacts and ensure data consistency. 
To enhance practicality and usability, the system replaces manual patient diaries with automated non-wear episode detection and sleep segmentation, reducing reliance on potentially subjective reporting. 
Beyond RBD detection, this module can be used as a standalone tool for general-purpose actigraphy analysis.\\
The ML module comprises two sequential components. First, a feature‑extraction stage distills each subject’s sleep‑motion patterns into a set of interpretable numerical descriptors, developed in collaboration with a sleep expert (\cref{fig:pipeline_overview}b). Next, a boosted decision-tree classifier maps these features  onto nightly RBD probability scores (\cref{fig:pipeline_overview}c). To mitigate night‑to‑night variability, nightly scores are aggregated into a single patient‑level prediction.\\
Model development was based on a training cohort of 78 individuals, including 55 with iRBD and 23 healthy controls (HC), and was evaluated on an independent local test cohort of 31 participants (19\, iRBD, 12\,HC).
To assess generalizability, the model was validated on two external cohorts: one of 103 individuals (70\,iRBD, 8\,\pdrbd{}, 25\,HC) and another of 31 individuals (13\,iRBD, 10\,\pdnorbd{}, 3\,\pdrbd{}, and 3\,HC), with some participants contributing multiple recordings.
Here, \pdrbd{} and \pdnorbd{} denote PD participants with and without PSG-confirmed RBD, respectively.
To comprehensively assess robustness, we performed leave-one-dataset-out cross-validation across all cohorts.
Moreover, we conducted a limited exploratory cross-device evaluation under lower prevalence conditions using a small, single-night publicly available cohort of 28 participants (3\,RBD, 25\,non-RBD).

Collectively, these analyses demonstrate consistent performance across diverse populations, supporting the potential of our approach as a broadly applicable framework for actigraphy-based RBD detection.

\section{Results}\label{sec:results}
\subsection{Actigraphy Cohorts Overview}\label{subsec:cohort_overvie}
Four distinct datasets from three different countries were used in this study.  
Initial model development was conducted on training data from \textit{\cgntrain{}}, which comprised 78 individuals (55\,iRBD, 23\,HC) enrolled in the CogTrAiL-RBD randomized controlled trial\,\cite{ophey_cognitive_2024} in Germany, contributing a total of 524 recorded nights. \\
Our internal evaluation set (\textit{\cgntest{}}), included 31 prospectively recruited participants (19\,iRBD, 12\,HC) from an ongoing iRBD screening effort at the same site\,\cite{seger_evaluation_2023}. \\ 
External validation relied on two additional cohorts.
One of these was the \oxflong{} (hereafter referred to as \textit{\oxfshort{}} for brevity).
It comprised 103 individuals (70\,iRBD, 8\,\pdrbd{}, 25\,HC) with several participants contributing multiple dominant-hand recordings from longitudinal follow-up, yielding a total of 113 actigraphy samples.\\
The other external validation cohort included 31 individuals (13\,iRBD, 10\,\pdrbd{}, 5\,\pdnorbd{}, 3\,HC) recruited from ongoing cohorts at the Lundbeck Foundation Parkinson’s Disease Research Center (PACE), Denmark with multiple recordings per participant resulting in 57 samples. This cohort will hereafter be referred to as \textit{\aar{}}.
Across all cohorts, each recording typically spanned 6–7 consecutive nights, providing in total more than 1,809 nights of actigraphy for model development and validation.
All recordings were acquired using the Axivity AX6 device with standardized acquisition settings like sample rate ($100\,\text{Hz}$) and dynamic range  ($\pm8\,\text{g}$).
Further details of the study design and data acquisition protocols are provided in Section~\ref{subsec:study}.

An overview of demographic and recording characteristics across all cohorts is summarized in Table~\ref{table:datasets}.
Of the recorded variables, the RBD Screening Questionnaire (RBDSQ) and motor impairment (MDS-UPDRS-III) differed significantly between subgroups, consistent with the expected separation of RBD cases from healthy controls.
In the \oxfshort{} cohort, additional group differences were observed for cognition (MoCA) and olfaction (Sniffin’ Sticks, SSI), reflecting the known decline of these domains in Parkinsonian and RBD populations. No significant cognitive or olfactory differences were found in the \aar{} cohort, and corresponding values were partly unavailable for the \cgntrain{} and \cgntest{} datasets.
Medication prevalence across cohorts is summarized in Table S6, with potentially motor-influential medications present among the participants.

\noindent\begin{minipage}{\linewidth}
{\tiny 
\centering
\begin{threeparttable}
\captionof{table}{\textbf{Dataset demographics and characteristics.} Values are given either as record-level average over each subgroup ($\operatorname{mean}\pm\operatorname{SD}$) or as proportions.
Clinical indicators capture prodromal features of $\alpha$-synucleinopathies and include measures of sleep disturbance (RBDSQ), motor function (UPDRS-III), cognition (MoCA), and olfaction (SSI). 
Statistically significant differences $\left(p <0.05\right)$ are highlighted in bold.}
\label{table:datasets}
\setlength{\tabcolsep}{1pt}
\renewcommand{\arraystretch}{1.9}
\begin{tabular}{ll  cc cc ccc  ccc cccc}
\cmidrule{3-13}
\multirow{2}{*}{\textbf{}} & \multirow{2}{*}{\textbf{}} 
& \multicolumn{2}{c}{\cellcolor{TableColor2}\textbf{\cgntrain{}}} 
& \multicolumn{2}{c}{\cellcolor{TableColor2}\textbf{\cgntest{}}} 
& \multicolumn{3}{c}{\cellcolor{TableColor2}\textbf{\oxfshort{}}}
& \multicolumn{4}{c}{\cellcolor{TableColor2}\textbf{\aar{}}} \\
\cmidrule{3-13}
&  & \cellcolor{TableColor1}iRBD & \cellcolor{TableColor1}HC
   & \cellcolor{TableColor2}iRBD &\cellcolor{TableColor2} HC
   & \cellcolor{TableColor1}iRBD & \cellcolor{TableColor1}\pdrbd{} &\cellcolor{TableColor1} HC 
   & \cellcolor{TableColor2}iRBD & \cellcolor{TableColor2}\pdrbd{} & \cellcolor{TableColor2}\pdnorbd{}&\cellcolor{TableColor2} HC \\
\midrule
\textbf{Samples}\textsuperscript{a}
  & value & \cellcolor{TableColor1}55 & \cellcolor{TableColor1}23 
                             & \cellcolor{TableColor2}19 & \cellcolor{TableColor2}12 
                             & \cellcolor{TableColor1}80 & \cellcolor{TableColor1}8 & \cellcolor{TableColor1}25 
                             & \cellcolor{TableColor2}23 & \cellcolor{TableColor2}19 & \cellcolor{TableColor2}9 & \cellcolor{TableColor2}6\\                     
\midrule
\multirow{2}{*}{\textbf{Nights}} 
& value     & \cellcolor{TableColor1}$6.7\pm0.6$ & \cellcolor{TableColor1}$6.7\pm0.9$ 
            & \cellcolor{TableColor2}$6.3\pm1.2$ & \cellcolor{TableColor2}$6.6\pm0.8$ 
            & \cellcolor{TableColor1}$6.2\pm1.4$ & \cellcolor{TableColor1}$6.2\pm0.7$  & \cellcolor{TableColor1}$5.8\pm2.0$
             & \cellcolor{TableColor2}$7.4\pm1.4$ & \cellcolor{TableColor2}$6.2\pm1.2$  & \cellcolor{TableColor2}$6.3\pm1.5$ & \cellcolor{TableColor2}$6.3\pm1.9$\\
(per sample) & $p$ & \multicolumn{2}{c}{\cellcolor{TableColor1}\textsuperscript{i}0.467} 
              & \multicolumn{2}{c}{\cellcolor{TableColor2}\textsuperscript{i}0.333} 
              & \multicolumn{3}{c}{\cellcolor{TableColor1}\textsuperscript{ii}0.689}
              & \multicolumn{4}{c}{\cellcolor{TableColor2}\textsuperscript{ii}0.055} \\           
\midrule
\multirow{2}{*}{\textbf{Age}} 
  & value     & \cellcolor{TableColor1}$69.6\pm5.9$ & \cellcolor{TableColor1}$67.6\pm 4.0$ 
              & \cellcolor{TableColor2}$68.4\pm 5.0$ & \cellcolor{TableColor2}$69.8\pm 6.6$ 
              & \cellcolor{TableColor1} $70.6\pm 6.9$  & \cellcolor{TableColor1} $74.3\pm 5.6$ &  \cellcolor{TableColor1} $69.8\pm 8.9$ 
              & \cellcolor{TableColor2}$65.8\pm 5.9$ & \cellcolor{TableColor2}$68.4\pm 7.9$ & \cellcolor{TableColor2}$70.9\pm 7.0$ & \cellcolor{TableColor2}$51.4\pm 2.9$ \\
 (yrs) & $p$ & \multicolumn{2}{c}{\cellcolor{TableColor1}\textsuperscript{i}0.098} 
              & \multicolumn{2}{c}{\cellcolor{TableColor2}\textsuperscript{i}0.623} 
              & \multicolumn{3}{c}{\cellcolor{TableColor1}\textsuperscript{ii}0.282}
              & \multicolumn{4}{c}{\cellcolor{TableColor2}\textsuperscript{ii}\textbf{0.003}} \\          
\midrule
\multirow{2}{*}{\textbf{Sex}} 
& value     & \cellcolor{TableColor1}47/8 & \cellcolor{TableColor1}21/2 
            & \cellcolor{TableColor2}14/5 & \cellcolor{TableColor2}12/2 
            & \cellcolor{TableColor1} 75/5 & \cellcolor{TableColor1} 7/1 & \cellcolor{TableColor1} 10/15
            & \cellcolor{TableColor2} 19/4 & \cellcolor{TableColor2}14/5 & \cellcolor{TableColor2}7/2 & \cellcolor{TableColor2} 4/2\\
(m/f) & $p$ & \multicolumn{2}{c}{\cellcolor{TableColor1}\textsuperscript{iii}0.714} 
              & \multicolumn{2}{c}{\cellcolor{TableColor2}\textsuperscript{iii}0.670} 
              & \multicolumn{3}{c}{\cellcolor{TableColor1}\textsuperscript{iv} $\mathbf{1.2\times10^{-7}}$}
              & \multicolumn{4}{c}{\cellcolor{TableColor2}\textsuperscript{iv}0.841}\\
\midrule
\multirow{2}{*}{\textbf{RBDSQ}} 
& value     &\cellcolor{TableColor1} $8.8\pm3.1$ & \cellcolor{TableColor1}$2.3\pm2.0$
            &\cellcolor{TableColor2} $9.8\pm1.9$ & \cellcolor{TableColor2} {\tiny{na}}
            & \cellcolor{TableColor1} $9.4\pm 2.3$  & \cellcolor{TableColor1} $10.8\pm 2.7$& \cellcolor{TableColor1} $1.7\pm 1.9$
            & \cellcolor{TableColor2}$10.1\pm 1.8$ & \cellcolor{TableColor2}$7.8\pm 4.0$ & \cellcolor{TableColor2}$3.2\pm 1.5$ & \cellcolor{TableColor2}$0.5\pm 0.6$\\
& $p$& \multicolumn{2}{c}{\cellcolor{TableColor1}\textsuperscript{i}$\mathbf{2.7\times10^{-9}}$} 
              & \multicolumn{2}{c}{\cellcolor{TableColor2} {\tiny{na}}} 
              & \multicolumn{3}{c}{\cellcolor{TableColor1} \textsuperscript{ii}$\mathbf{1.6\times10^{-10}}$}
              & \multicolumn{4}{c}{\cellcolor{TableColor2}\textsuperscript{ii}$\mathbf{1.0\times10^{-5}}$}\\
\midrule
\multirow{2}{*}{\textbf{MDS-}} 
& value     & \cellcolor{TableColor1}$7.2\pm4.3$ & \cellcolor{TableColor1} $4.3\pm3.9$
            & \cellcolor{TableColor2}$5.6\pm3.6$ & \cellcolor{TableColor2} {\tiny{na}}
            &  \cellcolor{TableColor1} $12.4\pm 9.7$ & \cellcolor{TableColor1}  $33.0\pm 14.7$& \cellcolor{TableColor1} $5.4\pm 3.8$
            & \cellcolor{TableColor2}$8.9\pm7.5$ & \cellcolor{TableColor2}$25.1\pm 11.2$ & \cellcolor{TableColor2}$29.6\pm 16.8$ & \cellcolor{TableColor2}$0.0\pm 0.0$ \\
\textbf{UPDRS-III} & $p$& \multicolumn{2}{c}{\cellcolor{TableColor1} \textbf{\textsuperscript{i}0.001}} 
              & \multicolumn{2}{c}{\cellcolor{TableColor2} {\tiny{na}}} 
              & \multicolumn{3}{c}{\cellcolor{TableColor1} \textsuperscript{ii}$\mathbf{2.0\times10^{-6}}$}
              & \multicolumn{4}{c}{\cellcolor{TableColor2}\textsuperscript{ii}$\mathbf{4.4\times10^{-7}}$}\\
\midrule
\multirow{2}{*}{\textbf{MoCA}} 
& value     & \cellcolor{TableColor1}$26.7\pm2.5$ & \cellcolor{TableColor1} $27.2\pm1.9$
            & \cellcolor{TableColor2}$26.7\pm1.6$ & \cellcolor{TableColor2} {\tiny{na}}
            & \cellcolor{TableColor1}$25.7\pm2.8$ & \cellcolor{TableColor1} $23.4\pm5.6$& \cellcolor{TableColor1}$27.2\pm2.2$
            & \cellcolor{TableColor2}$27.3\pm 2.3$ & \cellcolor{TableColor2}$27.7\pm 1.9$ & \cellcolor{TableColor2}$28.2\pm 1.7$ & \cellcolor{TableColor2}$29.0\pm 1.2$\\
& $p$& \multicolumn{2}{c}{\cellcolor{TableColor1}\textsuperscript{i}0.682} 
              & \multicolumn{2}{c}{\cellcolor{TableColor2}{\tiny{na}}} 
              & \multicolumn{3}{c}{\cellcolor{TableColor1}\textsuperscript{ii}\textbf{0.039}} 
              & \multicolumn{4}{c}{\cellcolor{TableColor2}\textsuperscript{ii}0.473}\\
\midrule
\multirow{2}{*}{\textbf{SSI}} 
& value     & \cellcolor{TableColor1}$6.5\pm2.6$ & \cellcolor{TableColor1}{\tiny{na}}  
            & \cellcolor{TableColor2}$7.3\pm2.1$ & \cellcolor{TableColor2}{\tiny{na}}   
            &  \cellcolor{TableColor1} $7.3\pm3.3$& \cellcolor{TableColor1} $5.0\pm2.5$& \cellcolor{TableColor1} $12.3\pm0.6$
            & \cellcolor{TableColor2}$7.1\pm 3.3$ & \cellcolor{TableColor2}$6.7\pm 2.7$ & \cellcolor{TableColor2}$7.8\pm 2.2$ & \cellcolor{TableColor2}$14.0\pm 0.1$\\
& $p$& \multicolumn{2}{c}{\cellcolor{TableColor1}{\tiny{na}}} 
              & \multicolumn{2}{c}{\cellcolor{TableColor2}{\tiny{na}}} 
              & \multicolumn{3}{c}{\cellcolor{TableColor1}\textsuperscript{ii}\textbf{0.012}} 
              & \multicolumn{4}{c}{\cellcolor{TableColor2}\textsuperscript{ii}0.068}\\
\bottomrule
\end{tabular}
\vspace{0.5em}

\begin{tablenotes}[flushleft]
\footnotesize
\item   \textbf{RBDSQ}: REM Sleep Behavior Disorder Screening Questionnaire (0-13 points; scores above 5-6 suggest probable RBD)\,\cite{stiasny-kolster_rem_2007};
        \textbf{MDS-UPDRS-III}: Movement Disorder Society–Unified Parkinson’s Disease Rating Scale, Part\,III (0–132 points; below 32 is mild, above 59 is severe)\,\cite{goetz_movement_2008,skorvanek_differences_2017};
        \textbf{MoCA}: Montreal Cognitive Assessment (0–30 points; normal 26–30, mild impairment 18-25, moderate 10-17, severe 0-9)\,\cite{nasreddine_montreal_2005};
        \textbf{SSI}: Sniffin’ Sticks Identification test (0-16 points; 12-16 normal olfaction), 9-11 hyposmia, 0-8 anosmia\,\cite{hummel_sniffin_1997}.
        \textbf{na}: Data not available.
\item[\textbf{a}]   \textbf{Number of actigraphy recordings}: For \cgntrain{} and \cgntest{}, this equals the number of subjects. For \oxfshort{}, recordings from 103 individuals (70 iRBD, 8 \pdrbd{}, 25 HC), i.e. ten individuals contributed two records.
                    For \aar{}, recordings from 31 individuals (13 iRBD, 10 \pdrbd{}, 3 \pdnorbd{}, 3 HC) were included in the displayed test set, while an additional 32 recordings from 18 individuals (1 iRBD, 4 \pdrbd{}, 6 \pdnorbd{}, 7 HC) were used for training only, as they contained an insufficient number of nights $(1.2\pm0.5)$. The demographics of these training-only cases did not differ significantly from those of the test set within each subgroup (all $p > 0.05$), except for marginal differences in sex distribution in \pdrbd{} $(p = 0.05)$ and in RBDSQ/UPDRS-III scores in HC ($p = 0.03$ and $p = 0.04$, respectively).
\item[\textbf{i-iv}] \textbf{Statistical tests}: 
    \textsuperscript{\textbf{i}} two-sided Mann-Whitney U. 
    \textsuperscript{\textbf{ii}} Kruskall-Wallis.
    \textsuperscript{\textbf{iii}} Fisher-Irwin.
    \textsuperscript{\textbf{iv}} Fisher-Freeman-Halton.
\end{tablenotes}
\end{threeparttable}
}
\end{minipage}\\

\subsection{Generalizable RBD Models through Robust Preprocessing}\label{subsec:preprocessing_results}
Developing generalizable ML models for RBD screening from actigraphy data requires a preprocessing pipeline that can reliably harmonize data across heterogeneous input sources. 
We designed a robust signal preprocessing framework aimed at standardizing raw accelerometer recordings from different devices and cohorts, ensuring compatibility for downstream feature extraction and model inference.\\
The pipeline consists of resampling, bandpass filtering, non-wear episode detection, auto-calibration, and sleep–wake segmentation (see \cref{subsec:preprocessing}\,\&\,\cref{fig:pipeline_overview}). We focus on two key components of this process, demonstrating their impact on standardization and data quality.
First, we assess the effectiveness of device auto-calibration by quantifying deviations from the expected unit sphere before and after correction, following the method proposed by van Hees et al. (2014)\,\cite{van_hees_autocalibration_2014}.
Second, we evaluate the performance of automated sleep segmentation by comparing algorithmically derived sleep windows to manually annotated sleep diaries.

\begin{figure}[htbp]
	\centering
	\includegraphics[page=1, width=\textwidth]{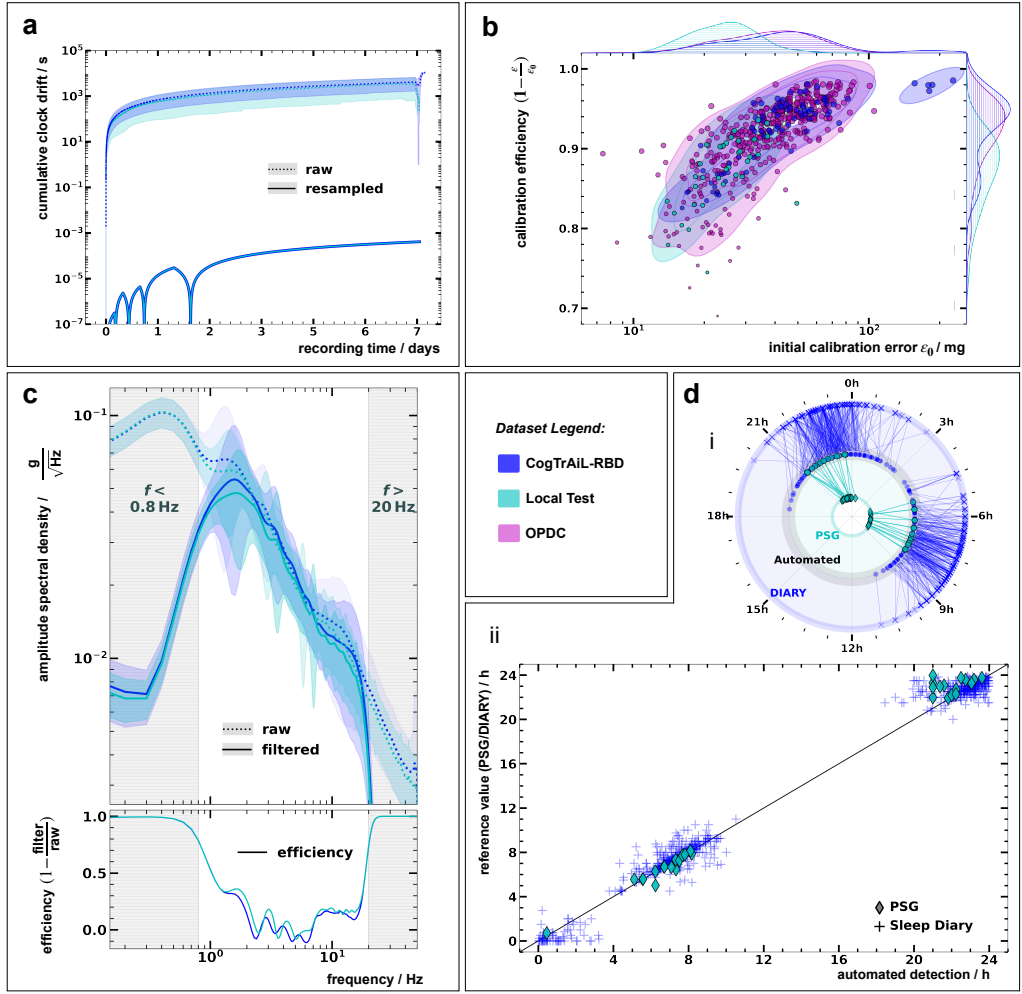}  
    \vspace{-.6cm}
	\caption{\label{fig:preprocessing_results}
    \textbf{Robust preprocessing for generalizable RBD detection.} Overview of preprocessing steps and validation; cohort colors match the legend (bottom center).
    {\textbf{(a) Resampling.} 
    Cumulative clock drift over recording time. Raw actigraphy signals sampled at a nominal $100\,\textrm{Hz}$ show substantial timing drift due to internal clock inaccuracies. Resampling corrects this drift to within numerical precision, as evidenced by near-identical post-resampling curves across cohorts.}
    {\textbf{(b) Calibration.} 
    Initial calibration error $\epsilon_0$ vs reduction efficiency $1 - \nicefrac{\epsilon}{\epsilon_0}$, where $\epsilon$ is the post-calibration error. Calibration is highly effective across cohorts, with $\operatorname{mean}\pm\operatorname{SD}\,[\operatorname{95\%CI}]$ efficiencies of $0.93\pm0.04\,[0.92,0.94]$ (\cgntrain{}), $0.87\pm0.04\,[0.85,0.89]$ (\cgntest{}{}), and $0.91\pm0.05\,[0.91,0.92]$ (\oxfshort{}{}). Higher initial errors yield greater correction gains.}
    {\textbf{(c) Filtering.}
    Amplitude spectral density (ASD) before (dotted line) and after (solid line) bandpass filtering, highlighting suppression of noise outside while preserving signal power within the $0.8\textrm{\textendash}20\,\textrm{Hz}$ passband. Cohort-averaged ASDs (Welch’s method) align closely outside the band but show greater variability (SD shown by shaded area) within it, supporting the choice of frequency cutoffs that isolate signal-dominated activity. Retention and suppression scores were $0.78\pm0.01\,[0.77, 0.78]\,/\,0.89\pm0.01\,[0.89, 0.90]$ for \cgntrain{} data, and $0.73\pm0.09 [0.70, 0.77]\,/\,0.90\pm0.01\,[0.89, 0.90]$ for \cgntest{} data.
    }
    {\textbf{(d) Sleep-Detection.}
    Comparison of automatically detected sleep onset and wake-up times with reference values from sleep diaries (\cgntrain{}, $n=756$) and PSG (\cgntest{}, $n=32$). Subfigure (i) displays predicted and reference times in clock format; the closer the connecting lines are to perfectly radial, the stronger the temporal alignment. Subfigure (ii) shows a scatter plot of automated versus reference times. Strong agreement is evidenced by Pearson correlation coefficients of $0.994\pm0.001\,[0.994,0.995]$ (\cgntrain{}), $0.996\pm0.001\,[0.992,0.998]$ (\cgntest{}{}) and mean-absolute errors (in minutes) of $34.4\pm40.9\,[31.5,37.3]$ minutes (\cgntrain{}), $35.8\pm45.5\,[19.4,52.3]$ (\cgntest{}{}). 
    The relatively large SDs compared to the means reflect some high-variance nights, while the narrow confidence intervals suggest that the mean error estimates remain robust at the group level.}}
\end{figure}

\newpar{Auto-calibration}
Actimeters typically require individual calibration to correct for device-specific variations in gain and offset, especially when accurate axis-orientation relationships are critical.
However, manual calibration is labor-intensive and thus rarely performed in practice.
Therefore, we incorporate a post-hoc auto-calibration step into our pipeline, leveraging the observation that the magnitude of the acceleration vector approximates the standard gravitational acceleration during rest periods.
This involves identifying low-activity segments, computing deviations from the unit sphere, and iteratively fitting gain and offset parameters via weighted least square regression to minimize discrepancies between observed data points and the theoretical unit sphere\,\cite{van_hees_autocalibration_2014}.\\
Since calibration error can be directly quantified by the deviation from the unit sphere, we evaluated the effectiveness of our auto-calibration by comparing these deviations before and after correction (\cref{fig:preprocessing_results}b), which illustrates the relationship between initial error and relative error reduction ratio across cohorts.
Reflecting these trends, calibration errors were reduced from $50.89\pm38.28\,\textrm{mg}$ to $2.81\pm0.81\,\textrm{mg}$ across the \cgntrain{} cohort and from $25.13\pm7.65\,\textrm{mg}$ to $3.09\pm1.12\,\textrm{mg}$ within the \cgntest{} cohort.
Within the \oxfshort{} cohort, we observed a reduction from $40.01\pm16.72\,\textrm{mg}$ to $2.83\pm1.01\,\textrm{mg}$.

\newpar{Automated Sleep Segmentation}[par:sleepsegmentation]
In ambulatory studies targeting RBD detection from actigraphy, concurrent polysomnography (PSG) for sleep-window determination is typically unavailable.
In those studies sleep periods are approximated using participant-completed sleep diaries\,\cite{brink-kjaer_ambulatory_2023}, which are potentially limited by subjectivity, precision in reported times, or missing data due to participant burden, and light sensor data\,\cite{raschella_actigraphy_2023}, reflecting time in bed rather than actual sleep.\\
As part of our preprocessing pipeline, we employ the HDCZA algorithm introduced by van Hees et al. (2018)\,\cite{van_hees_estimating_2018} to identify candidate sleep periods from raw actigraphy, as the algorithm has been shown to generalize well across populations and to outperform both machine learning-based and other heuristic approaches\,\cite{patterson_40_2023}.
To assess the performance of this algorithm within our specific setting, we conducted two validation analyses: one using annotated sleep diaries from the training cohort and another using PSG recordings from the test cohort, both available for a subset of individuals.

For the diary-based validation, we analyzed data from the \cgntrain{} cohort, comprising 61 individuals {\small$\left(44\,\textrm{iRBD},\,17\,\textrm{HC}\right)$} with available sleep diary annotations spanning 7 nights each.
We computed per-subject c-statistics (area under the ROC curve from binary sleep/wake labels, equivalent to the concordance index in the absence of probabilistic scores \cite{van_hees_estimating_2018}) from binarized sleep/wake labels at 30-second resolution and averaged them across the cohort, achieving a mean c-statistic of \ciTwo{0.93}{0.91}{0.94}{0.01}.
Sleep onset and wake-up timestamps pooled across all individuals yielded a mean absolute error (MAE) of \ciOne{34.4}{31.5}{37.7}{0} minutes and a Pearson correlation of \ciThree{0.994}{0.993}{0.995}{0} as displayed in \cref{fig:preprocessing_results}d.\\
We further validated sleep detection performance using parallel PSG and actigraphy recordings from 16 participants {\small$\left(13,\textrm{iRBD},\,3,\textrm{HC}\right)$} of the \cgntest{} cohort, where PSG was available for the first night of each seven-day recording and sleep intervals were scored by an expert (M.S.).
The algorithm was applied to full seven-day recordings, with performance metrics derived from the first (PSG-recorded) night, yielding a mean c-statistic of \ciTwo{0.91}{0.87}{0.94}{0}, a mean absolute error of \ciOne{35.8}{19.4}{52.2}{0} minutes, and a Pearson correlation of \ciThree{0.996}{0.992}{0.998}{0}. 
Taken together, these results validate the HDCZA algorithm as a reliable component of our preprocessing pipeline, demonstrating strong agreement with both diary-based annotations and expert-scored PSG intervals. 
Notably, the automated segmentation shows a slight tendency to underestimate sleep duration relative to PSG (\cref{fig:preprocessing_results}d), which may be beneficial in our context by reducing the risk of calculation sleep features on the wake-period activity.
Although direct comparisons between the diary- and PSG-based evaluations are limited by differences in sample size, previous studies indicate that sleep diaries often overestimate sleep intervals relative to PSG\,\cite{matthews_similarities_2018, chou_comparison_2020, lehrer_comparing_2022}, suggesting that actigraphy-based approaches like HDCZA may offer a more objective alternative to diaries.

Together with robust resampling to correct sampling drift (\cref{fig:preprocessing_results}a) and effective noise reduction through bandpass filtering (\cref{fig:preprocessing_results}c), these steps ensure standardized and physiologically meaningful input signals, crucial for learning generalizable features in downstream machine learning models.

\subsection{Single-Center Model Generalizability}\label{subsec:ml_results}
Prior to developing our own model, we assessed a previously published actigraphy-based RBD detection approach\,\cite{raschella_actigraphy_2023} on our multi-centre data.
Its evaluation indicated limited cross-cohort generalizability (Table S1), motivating the development of a dedicated, robust pipeline.\\
Leveraging the standardized, noise-reduced actigraphy signals generated by our preprocessing module (\cref{fig:preprocessing_results}), we developed and prototyped a single-centre machine-learning model on the \cgntrain{} cohort to assess its transferability to cohorts originating from different centers.

\begin{figure}[htbp] 
	\centering
	\includegraphics[page=1, width=\textwidth]{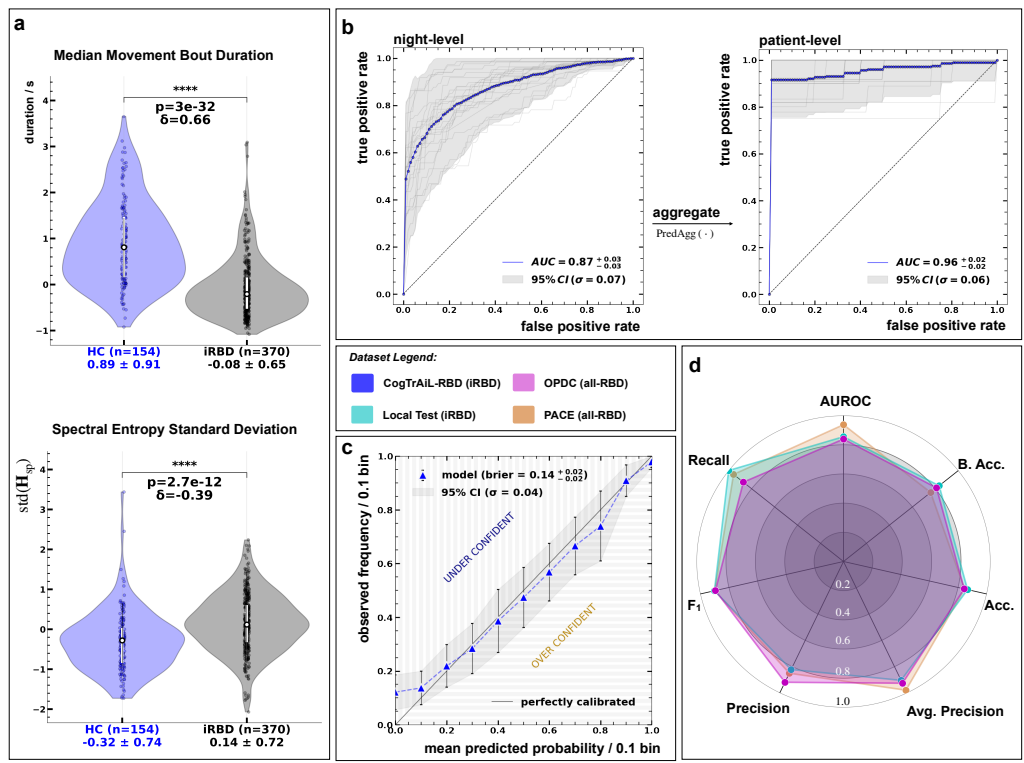}
    \vspace{-.4cm}
	\caption{\label{fig:model_results}
    \textbf{Predictive RBD Modeling Results.}
    \textbf{(a)} Violin plots of two selected features illustrating distributional shifts between individuals with RBD and healthy controls. P-values are computed using two-sided Mann–Whitney U tests, and effect sizes $\left(\delta\right)$ are reported as Cliff’s delta. These features are discussed in more detail at the end of the results section. 
    \textbf{(b)} ROC curves of the nested cross-validation results of the night-level prediction (left) and after aggregation to the patient level (right). The blue line indicates the mean over all folds, and the shaded area represents the 95\% confidence interval. The improved performance after aggregation reflects the benefit of multi-night actigraphy and helps mitigate night-to-night variability in motor activity.
    \textbf{(c)} Calibration curve on the night level using predictions from nested cross-validation. Triangles indicate the observed positive rate per probability bin; the shaded region shows the 95\% CI across folds. The predicted probabilities are well calibrated and closely reflect the true likelihood of RBD.
    \textbf{(d)} Radar plot summarizing classifier performance across multiple evaluation metrics for the external test sets. Results are shown separately for the \cgntest{} cohort (cyan), the \oxfshort{} cohort (magenta) and the \aar{} cohort (dark-orange), where (iRBD) and (\allrbd{}) denote the respective classification tasks (see \cref{table:validation_results}), indicating robust and balanced generalization with a subtle emphasis on recall.}
\end{figure}

\newpar{Single-Center Model Development and Internal Validation}[par:CGNtrain-results]
After data preprocessing we extract numerical features characterizing motor behavior during sleep. 
Engineered with the aim of differentiating between RBD and non-RBD individuals, these features leverage domain-specific knowledge and clinical experience, incorporating knowledge from a trained RBD expert (M.S.).
Specifically, they capture distinct patterns within individuals’ movements during sleep, including intensity, periodicity, spectral properties distinguishing rapid from slow movements, complexity, fragmentation, overall activity levels, and clustering behavior, i.e., whether movement bursts are uniformly distributed throughout the night or occur in temporal clusters, for example during REM periods.
For a detailed description and physiological interpretation of each feature, see Table S2.
The ML model was developed and trained exclusively on data from the \cgntrain{} cohort (78\,individuals,\,524\,nights), with feature selection and hyperparameter tuning performed within a nested cross-validation framework to prevent biased performance estimates, as detailed in \cref{subsec:methods_model_developments}.\\
Classification performance, assessed on the outer folds of the nested cross-validation, achieved a mean area under the receiver operating characteristic curve (AUROC) of \ciTwo{0.87}{0.84}{0.90}{0.07}, a $F_1$ score of \ciTwo{0.82}{0.79}{0.85}{0.06} and balanced accuracy of \ciTwo{0.81}{0.78}{0.84}{0.07} for the night-level prediction.
On the patient level, after aggregating over all available nights of each patient ($6.4\pm0.9$), the classification reaches a mean AUROC of \ciTwo{0.95626}{0.93297}{0.97955}{0.05642}, a $F_1$ score of \ciTwo{0.92492}{0.89796}{0.95189}{0.06532} and balanced accuracy of \ciTwo{0.92333}{0.88788}{0.95878}{0.08589}, demonstrating the benefit of collecting actigraphy data over multiple nights to mitigate night-to-night variability (see \cref{fig:model_results}b).
Calibration of non-thresholded probability predictions was assessed using the Brier score, yielding values of \ciTwo{0.14}{0.13}{0.16}{0.04} at the night level and \ciTwo{0.12}{0.09}{0.13}{0.05} at the patient level, suggesting that the predicted probabilities reasonably reflect the true RBD likelihood as displayed in \cref{fig:model_results}c.

\begin{table}[htbp]
\centering
\caption{\textbf{Single-Center Model Validation Results.}
Shown are the datasets used to validate the ML model, along with corresponding patient-level classification metrics as mean with $95\%$ confidence intervals. Cohorts labeled iRBD were evaluated on the task of distinguishing iRBD from healthy controls (HC), while \pdrbd{} cohorts were used to differentiate PD individuals with PSG-confirmed RBD from HCs. The records column reports the number of multi-day actigraphy recordings, with the number of nights per class shown in parentheses.}
\setlength{\tabcolsep}{.2cm}
\renewcommand{\arraystretch}{1.6}
\rowcolors{2}{TableColor2}{TableColor1}
\begin{tabular}{>{\raggedleft}p{2.1cm}  >{\raggedleft}p{2.8cm} r c c c c}
\toprule
\textbf{Cohort}  &\textbf{Records} & \textbf{Model(s)} &\textbf{AUROC} & $\mathbf{F_1}$ & \textbf{Bal. Acc.} \\

\midrule 
\cellcolor{TableColor2}\textbf{\cgntrain{}} (\textbf{iRBD})  & \mbox{}\hspace{2.1em}iRBD: \phantom{0}55 (370) \newline HC: \phantom{0}23 (154)  &  CV & 
 \ciTwo{0.95626}{0.93297}{0.97955}{0.05642} &  \ciTwo{0.92492}{0.89796}{0.95189}{0.06532} &  \ciTwo{0.92333}{0.88788}{0.95878}{0.08589}\\
 
\midrule
\cellcolor{TableColor2}\textbf{\cgntest{}} (\textbf{iRBD}) & \mbox{}\hspace{2.1em}iRBD: \phantom{0}19 (119) \newline HC: \phantom{0}12 (\phantom{0}79) & Held-out &
 \ciThree{0.85533}{0.85218}{0.85848}{0.07186} &  \ciThree{0.90115}{0.89954}{0.90277}{0.03685} &  \ciThree{0.83267}{0.82966}{0.83567}{0.06852}\\
\midrule
\multirow[t]{2}{*}{\cellcolor{TableColor2}\textbf{\oxfshort{}}} (\textbf{iRBD}) & \mbox{}\hspace{2.1em}iRBD: \phantom{0}80 (496) \newline HC: \phantom{0}25 (144) & Held-out &
 \ciThree{0.83772}{0.83527}{0.84017}{0.05590} &  \ciThree{0.89026}{0.88907}{0.89145}{0.02714} &  \ciThree{0.80999}{0.80786}{0.81211}{0.04852}\\

\specialrule{0.04em}{0pt}{0pt}
\multirow[t]{2}{*}{\cellcolor{TableColor2}\textbf{\oxfshort{}}} (\textbf{\pdrbd{}}) & \mbox{}\hspace{.0em}\pdrbd{}: \phantom{00}8 (\phantom{0}50) \newline HC: \phantom{0}25 (144) & Held-out &  \ciThree{0.84363}{0.84056}{0.84670}{0.06993} &  \ciThree{0.67124}{0.66729}{0.67519}{0.09012} &  \ciThree{0.81309}{0.80991}{0.81627}{0.07252} \\

\specialrule{0.04em}{0pt}{0pt}
\multirow[t]{3}{*}{\cellcolor{TableColor2}\textbf{\oxfshort{}}} (\textbf{\allrbd{}}) & \mbox{}\hspace{2.1em}iRBD: \phantom{0}80 (496) \hspace{.4em}\pdrbd{}: \phantom{00}8 (\phantom{0}50) \newline HC: \phantom{0}25 (144) & Held-out &  \ciThree{0.83925}{0.83686}{0.84165}{0.05466} & \ciThree{0.89424}{0.89313}{0.89536}{0.02542} & \ciThree{0.80976}{0.80764}{0.81188}{0.04835} \\

\midrule
\multirow[t]{2}{*}{\cellcolor{TableColor2}\textbf{\aar{}}}  \mbox{}\hspace{2.1em}({\textbf{iRBD}}) & \mbox{}\hspace{2.1em}iRBD: \phantom{0}23 (171) \newline HC: \phantom{00}6 (\phantom{0}57)& Held-out & \ciThree{0.97276}{0.97098}{0.97453}{0.04054} & \ciThree{0.95741}{0.95556}{0.95925}{0.04209} & \ciThree{0.96065}{0.95901}{0.96229}{0.03739}
\\

\specialrule{0.04em}{0pt}{0pt}
\multirow[t]{3}{*}{\cellcolor{TableColor2}\textbf{\aar{}}} ({\textbf{\pdrbd{}{}}}) & \pdrbd{}: \phantom{0}19 (117) \newline \mbox{}\hspace{.3em}\pdnorbd{}: \phantom{00}9 (\phantom{0}38)\newline HC: \phantom{00}6 (\phantom{0}57)& Held-out &
\ciThree{0.95632}{0.95440}{0.95824}{0.04382} & \ciThree{0.83591}{0.83357}{0.83825}{0.05338} & \ciThree{0.77650}{0.77278}{0.78022}{0.08480}\\

\specialrule{0.04em}{0pt}{0pt}
\multirow[t]{4}{*}{\cellcolor{TableColor2}\textbf{\aar{}}} ({\textbf{\allrbd{}}}) & \mbox{}\hspace{2.1em}iRBD: \phantom{0}23 (171) \newline \pdrbd{}: \phantom{0}19 (117) \newline \mbox{}\hspace{.3em}\pdnorbd{}: \phantom{00}9 (\phantom{0}38)\newline HC: \phantom{00}6 (\phantom{0}57)& Held-out &
 \ciThree{0.93766}{0.93576}{0.93955}{0.04321} &  \ciThree{0.89953}{0.89794}{0.90112}{0.03625} &  \ciThree{0.75884}{0.75507}{0.76261}{0.08597}\\

\bottomrule
\end{tabular}
\footnotesize{
\textbf{CV}: Metrics reflect the mean across cross-validation folds, with $95\%$ confidence intervals derived from inter-fold variability, capturing both data split and model training variation.\\
\textbf{Held-out}: A single pretrained model is evaluated once on a blinded dataset. Confidence intervals ($95\%$) are estimated via bootstrap resampling 
($n=2000$, stratified by class) and quantify the uncertainty of the point estimate given the finite test set. 
}
\label{table:validation_results}
\end{table}

\newpar{Blinded Test Set and External Validation}[par:test-results]
While nested cross-validation offers a reliable assessment of internal performance, it does not fully capture the model’s ability to generalize to an independent, potentially out-of-distribution datasets. 
Hence, we trained a final model on the entire \cgntrain{} cohort and evaluated it on three independent datasets: a blinded holdout set from a prospective screening effort (\cgntest{}) and two external validation cohorts from different centers (\oxfshort{} and \aar{}), providing a stronger test of generalizability of our approach.
An overview of the model’s classification performance on the external test cohorts is shown in \cref{fig:model_results}d, while detailed metrics with 95\% confidence intervals—derived by bootstrap resampling for single-model evaluations and from inter-fold variability for cross-validation—are provided in Table~\ref{table:validation_results}.\\
Evaluation of the final model on the blinded \textit{\cgntest{}} cohort (31 individuals, 198 nights) yielded patient-level metrics of AUROC 0.86, $F_1$ 0.90, and balanced accuracy 0.83, indicating a moderate generalization gap between training and unseen data, as the AUROC falls below the lower bound of the internal 95\% confidence interval (0.94–0.98) from nested cross-validation.
Nonetheless, overall generalization remains strong, with classification metrics well balanced and a notably high $F_1$ score (0.90) reflecting strong sensitivity and slightly higher recall (0.99) than precision (0.83).
This asymmetry suggests a favorable bias toward detecting true positives, a desirable property in clinical screening where false negatives carry greater cost than false positives.
Results on the \oxfshort{} and \aar{} datasets (see Table~\ref{table:validation_results}) further support the model’s ability to generalize across clinical populations and recording conditions.\\
In the \oxfshort{} cohort, the model maintained strong generalization, achieving AUROC 0.84, $F_1$ 0.89, and balanced accuracy 0.81 for iRBD vs HC. Performance in PD+RBD vs HC was similar (AUROC 0.84) but with a lower $F_1$ (0.67) due to the small number of positive cases and resulting class imbalance.
Metrics for the combined iRBD and PD+RBD group (\allrbd{}) are predominantly influenced by the iRBD subset.\\
In the \aar{} cohort, the model achieved very strong iRBD detection with AUROC 0.97, $F_1$ 0.96, and balanced accuracy 0.96.
When \pdrbd{} cases were evaluated against HC/\pdnorbd{}, AUROC and $F_1$ remained high (0.96 and 0.84), but balanced accuracy declined to 0.78, indicating that greater clinical heterogeneity and class imbalance shift the optimal decision threshold despite preserved overall separability.

\newpar{Effect of Wrist Placement on Model Performance}[par:handedness-results]
To assess whether wrist dominance influences model performance, we evaluated 111 participants in the OPDC cohort with simultaneous dominant and non-dominant wrist recordings, where parallel bilateral recordings were available.
Using a paired permutation test (50\,000 permutations), we observed no statistically significant differences in performance between wrists (AUROC: dominant 0.85 vs. non-dominant 0.88, $p = 0.28$; Bal. Acc.: dominant 0.82 vs. non-dominant 0.81, $p = 0.84$; $F_1$: dominant 0.91 vs. non-dominant 0.90, $p = 0.62$), indicating that \algoname{}’s performance was not meaningfully affected by wrist placement in this cohort.

\newpar{Comparison to a Pretrained Self-Supervised Feature Extractor}[par:asleepres]
To contextualize both our handcrafted feature design and the sleep-detection component of the preprocessing pipeline, we additionally evaluated sleep detection, sleep staging, and RBD classification performance using a publicly available self-supervised sleep-staging framework (see \cref{subsec:model_design}).\,\cite{yuanSelfsupervisedLearningAccelerometer2024}\\
First, sleep–wake detection was evaluated under identical conditions as in \cref{subsec:preprocessing_results}. The model achieved mean absolute errors of \ciOne{51.49}{46.72}{56.07}{64.21} minutes (diary) and \ciOne{53.284}{36.414}{70.154}{46.791} minutes (PSG), consistently exceeding the error observed with our preprocessing pipeline ($\sim 35,\textrm{min}$).
Sleep staging was further assessed on the PSG subset by comparing predicted 30-second epoch labels to expert annotations (M.S.). Balanced accuracy reached \ciTwo{0.58}{0.52}{0.64}{0.11}, with a REM sensitivity of \ciTwo{0.209}{0.097}{0.321}{0.210}. Most REM epochs were misclassified as nREM.\\
Second, RBD classification was evaluated by replacing \algoname{}’s handcrafted features with the model’s 1024-dimensional latent representations.
Under identical nested cross-validation within the \cgntrain{} cohort, this approach achieved AUROC \ciTwo{0.926}{0.900}{0.952}{0.026}, F1 \ciTwo{0.888}{0.867}{0.910}{0.052}, and balanced accuracy \ciTwo{0.791}{0.759}{0.823}{0.032}.
On the independent \cgntest{} cohort, performance reached AUROC \ciTwo{0.849}{0.846}{0.852}{0.003}, F1 \ciTwo{0.808}{0.805}{0.811}{0.003}, and balanced accuracy \ciTwo{0.768}{0.764}{0.771}{0.003}.
These findings confirm that RBD-relevant signal is present in the high-dimensional self-supervised movement representations. However, performance remained consistently below that achieved with \algoname{}’s handcrafted features (\cref{table:validation_results}).
While the self-supervised features might increase representational richness, they may lack task-specific inductive bias for RBD and might fail to capture movement patterns most relevant to the disorder, particularly since RBD was not present in the training data of the feature extractor.

\subsection{Unified Multi-Center Model}\label{subsec:ml_results_pooled}
Having established that our single-centre prototype retains predictive power across external sites, we next pooled data from multiple cohorts to train a unified multi-centre model, intended as a generalizable, pre-trained solution for use in independent clinical settings. Training across sites aims to improve robustness by exposing the model to diverse populations and recording conditions.
To re-evaluate its generalization performance under realistic deployment scenarios, we employed leave-one-dataset-out (LODO) cross-validation, simulating application to previously unseen centers.
Demonstrating stability across these LODO models confirms that our model selection pipeline converges on consistent and meaningful solutions.
This ensures that the final pooled model—released publicly as a pre-trained resource—closely reflects the behaviour of the LODO models, and that LODO performance metrics provide realistic estimates of its expected performance on new cohorts.

\begin{figure}[htbp]
	\centering
	\includegraphics[page=1, width=\textwidth]{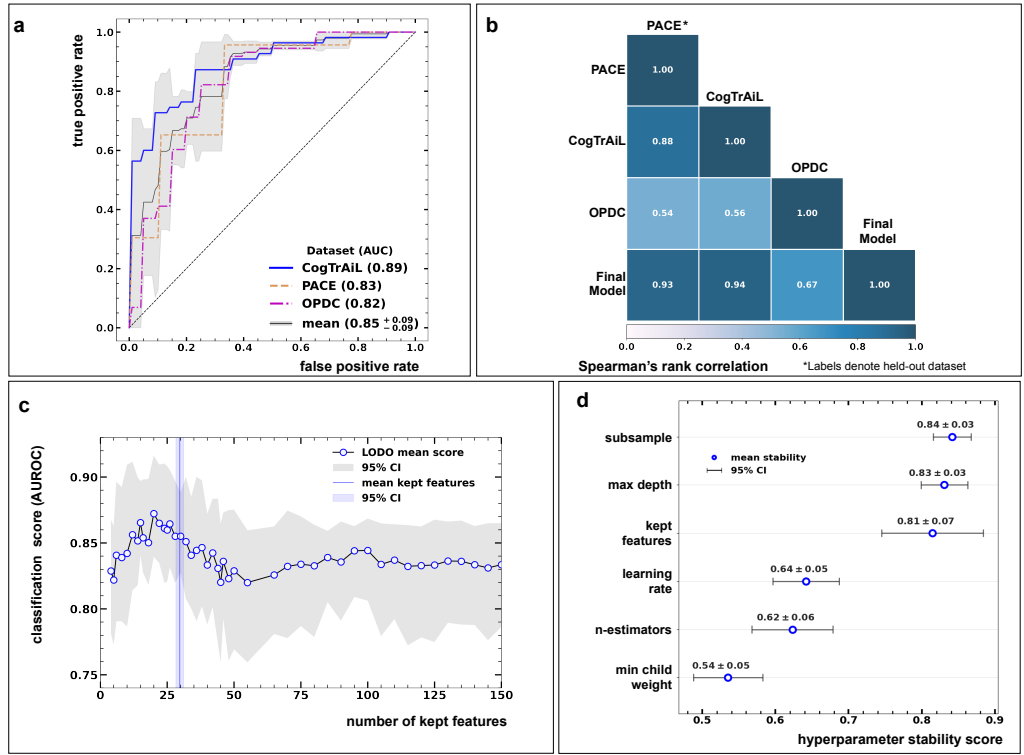} 
    \vspace{-.6cm}
	\caption{\label{fig:lodo_results}
    \textbf{Unified Multi-Center Model: Cross-Cohort Performance and Model Stability.}
    \textbf{(a) LODO Performance.} 
        ROC curves of the leave-one-dataset-out (LODO) cross-validation.
        Each curve corresponds to one fold, with one dataset held out for testing while the others were used for training.
        The results show consistently high discrimination across datasets, indicating robust generalization.
    \textbf{b) Feature Ranking Stability.} 
       Spearman’s rank correlation of \algoname{}’s inherent feature rankings across LODO folds, indicating consistently strong agreement (moderate-to-strong for OPDC holdout) and supporting the robustness of the final model.
    \textbf{(c) Feature Selection Stability \& Ablation} 
    Model performance as a function of features retained from a consensus ranking. Performance peaks around $\sim20$ features (the stable “core”), while the mean selected count across 20 seeded runs is slightly higher \ciTwo{29.738}{28.316}{31.159}{6.388} with a narrow band, indicating stable selection. Beyond this range, performance saturates and remains stable.
    \textbf{(d) Hyperparameter Stability.}
    Stability scores from repeated LODO runs (n=20) show that nearly all hyperparameters are highly stable, with only minor variability in a subset of hyperparameters. Overall, the training procedure converges to consistent configurations across cohorts, underscoring the robustness of the pipeline.
    }
\end{figure}

\newpar{Multi-Center Classification Performance}[par:lodo-results]
The multi-center model demonstrated strong overall generalization, confirming the robustness of our pipeline across independent cohorts.
Compared with the single-center model, it achieved similar discrimination (AUROC) but showed mixed F1-scores and slightly lower balanced accuracy (BA) on the same external datasets (see \cref{table:validation_results} and \cref{table:lodo_results}).
We noticed that including the \cgntest{} cohort in pooled training consistently preserved AUROC but lowered BA and increased Brier scores, indicating poorer probability calibration (Table S4). While the performance difference is minor, we excluded \cgntest{} from the final LODO training.\\
The performance differences among cohorts appear to stem mainly from calibration and regularization effects.
This argument is supported by higher Brier scores (e.g., \oxfshort{} \allrbd{} 0.12 vs 0.09, \aar{} \allrbd{} 0.14 vs 0.09, \oxfshort{} iRBD 0.12 vs 0.08, \aar{} iRBD 0.09 vs 0.06) for the multi-center model across cohorts, signaling less well calibrated probability scores, leading to modest BA and F1 reductions even when AUROC remains stable, as further elaborated in the Discussion.

Both the single-center and multi-center models are available in our GitHub repository (\giturl{}) so that users can explore different deployment scenarios and performance trade-offs.
\begin{table}[htbp]
\centering
\caption{\textbf{Multi-Center Model Validation Results.}
Displayed are classification metrics of the multi-center model across LODO folds, i.e. independent datasets. 
The confidence intervals ($95\%$) are estimated via bootstrap resampling 
($n=2000$, stratified by class). 
The LODO evaluation was performed twice—once on all RBD cases and once restricted to iRBD—yielding the upper and lower table blocks, respectively.}
\setlength{\tabcolsep}{.2cm}
\renewcommand{\arraystretch}{1.6}
\rowcolors{2}{TableColor2}{TableColor1}
\begin{tabular}{>{\raggedleft}p{2.5cm}  >{\raggedleft}p{2.8cm} r c c c}
\toprule
\textbf{Cohort}  &\textbf{Records} &\textbf{AUROC} & $\mathbf{F_1}$ & \textbf{Bal. Acc.} \\
\toprule 
\multirow[t]{3}{*}{\cellcolor{TableColor2}\textbf{\cgntrain{}}} ({\textbf{iRBD}}) & \mbox{}\hspace{2.1em}iRBD: \phantom{0}55 (370) \newline HC: \phantom{0}23 (154) & \ciThree{0.88787}{0.88621}{0.88953}{0.03789} & \ciThree{0.82442}{0.8226}{0.82624}{0.04149} & \ciThree{0.81913}{0.81722}{0.82103}{0.0435} \\
\specialrule{0.04em}{0pt}{0pt}
\multirow[t]{3}{*}{\cellcolor{TableColor2}\textbf{\oxfshort{}}} ({\textbf{\allrbd{}}}) & \mbox{}\hspace{2.1em}iRBD: \phantom{0}80 (496) \hspace{.4em}\pdrbd{}: \phantom{00}8 (\phantom{0}50) \newline HC: \phantom{0}25 (144) 
& \ciThree{0.82291}{0.82027}{0.82555}{0.06025} & \ciThree{0.91506}{0.91427}{0.91585}{0.01802} & \ciThree{0.73054}{0.72804}{0.73304}{0.05702} \\ 
\specialrule{0.04em}{0pt}{0pt}
\multirow[t]{3}{*}{\cellcolor{TableColor2}\textbf{\aar{}}} \mbox{}\hspace{2.1em}({\textbf{\allrbd{}}}) & \mbox{}\hspace{2.1em}iRBD: \phantom{0}23 (171) \newline \pdrbd{}: \phantom{0}19 (117) \newline \mbox{}\hspace{.3em}\pdnorbd{}: \phantom{00}9 (\phantom{0}38)\newline HC: \phantom{00}6 (\phantom{0}57)
& \ciThree{0.82784}{0.82395}{0.83173}{0.0887} & \ciThree{0.91851}{0.91694}{0.92009}{0.0359} & \ciThree{0.8152}{0.8116}{0.8188}{0.08211} \\
\midrule

\multirow[t]{3}{*}{\cellcolor{TableColor2}\textbf{\cgntrain{}}} ({\textbf{iRBD}}) & \mbox{}\hspace{2.1em}iRBD: \phantom{0}55 (370) \newline HC: \phantom{0}23 (154) & \ciThree{0.86729}{0.86536}{0.86923}{0.04406} & \ciThree{0.76923}{0.76718}{0.77128}{0.04678} & \ciThree{0.74557}{0.74327}{0.74788}{0.05252} \\
\specialrule{0.04em}{0pt}{0pt}
\multirow[t]{3}{*}{\cellcolor{TableColor2}\textbf{\oxfshort{}}}  \mbox{}\hspace{2.1em}({\textbf{iRBD}}) & \mbox{}\hspace{2.1em}iRBD: \phantom{0}80 (496) \newline HC: \phantom{0}25 (144)  &  \ciThree{0.85337}{0.85111}{0.85562}{0.05144} & \ciThree{0.91024}{0.90928}{0.9112}{0.02188} & \ciThree{0.76878}{0.76628}{0.77129}{0.05706} \\ 
\specialrule{0.04em}{0pt}{0pt}
\multirow[t]{3}{*}{\cellcolor{TableColor2}\textbf{\aar{}}} \mbox{}\hspace{2.1em} ({\textbf{iRBD}}) & \mbox{}\hspace{2.1em}iRBD: \phantom{0}23 (171) \newline HC: \phantom{00}6 (\phantom{0}57)& \ciThree{0.97276}{0.97098}{0.97453}{0.04054} & \ciThree{0.95741}{0.95556}{0.95925}{0.04209} & \ciThree{0.96065}{0.95901}{0.96229}{0.03739} \\

\bottomrule
\end{tabular}
\footnotesize{}
\label{table:lodo_results}
\end{table}

\subsection{Feature Selection and Hyperparameter Stability}
The model selection pipeline of \algoname{} incorporates feature selection and hyperparameter optimization.
To assess robustness, all engineered features were ranked by importance (see \cref{subsec:methods_model_developments}) and their consistency evaluated across the LODO folds. 
Spearman’s rank correlations were generally high (up to 0.94), confirming that the ranking procedure is internally stable under resampling.
Correlations were somewhat lower when OPDC was used as the held-out test set (0.54–0.67), yet they still indicate moderately strong agreement, which is expected given that OPDC is the largest cohort and its omission induces the greatest distributional shift (\cref{fig:lodo_results}b).
When comparing rankings derived within individual datasets, correlations were lower (0.36–0.70; Figure S1), reflecting cohort-specific feature preferences and underscoring the value of pooling data to derive the most generalizable feature set.

Next, we assessed feature selection stability.
For each fold of the LODO cross-validation, the pipeline selects a set of top-ranked features, and we quantified their overlap using the Jaccard similarity index.
We observed only a moderate overlap of \ciTwo{0.408}{0.391}{0.750}{0.1} (95\% CI). 
However, a stable core signal of ten features was consistently selected in all four CV folds, with seven of them also retained in the final pooled model. In addition, eleven supporting features appeared in two out of three folds, about half of which were also present in the final model. Thus, the modest Jaccard score largely reflects variability in these supporting features, while the stable core indicates reproducible structure across folds.
This raised the question of whether the stable core set of features alone would suffice to build accurate classification models.
To investigate this, we conducted an ablation study, re-training models while progressively increasing the number of retained features from a consensus ranking (\cref{fig:lodo_results}c).
Performance was already strong with the stable core features, and appeared to peak around 20–25 features — corresponding to the core plus a subset of supporting features.
The automatically selected feature sets performed slightly above the stable core alone and close to this peak, with a mean selected count of about 30 features across repeated LODO runs, suggesting that while the core features are highly informative, adding a limited number of supporting features yields the most reliable performance.
Based on these findings, we retained the pipeline’s automatic feature selection strategy for the final model, rather than restricting it to the core set of features alone.

Finally, we evaluated the stability of the hyperparameter optimization to assess whether tuning converged to similar configurations across folds.
Convergence would indicate that the model class and its regularization are robust to dataset composition, whereas divergence would suggest fold-specific sensitivity that could undermine generalization.
We rerun the LODO cross-validation 20 times under variation of the random seed and evaluated the population of hyperparameters in terms of a stability score (see \cref{subsec:methods_model_developments}).
We observed that roughly half of the hyperparameters, including the number of selected features, achieved high stability scores above $0.80$, while the remainder were somewhat lower ($0.54$–$0.64$) yet still in a moderate-to-strong range (see \cref{fig:lodo_results}d).
This indicates that, overall, the optimization converged consistently across repetitions, with only limited variability in some parameters.
Notably, the less stable ones often interact closely, such as the learning rate, number of trees (\texttt{n\_estimators}), and minimum child weight (\texttt{min\_child\_weight}, a parameter controlling tree complexity by enforcing a minimum leaf size). These parameters form compensatory trade-offs — for instance, smaller learning rates typically require more trees, and stricter child weight constraints can be balanced by deeper or more numerous trees. As a result, different but functionally equivalent configurations can yield comparable performance, leading to apparent variability without undermining robustness.

Taken together, the stability of feature rankings, selected feature sets, and hyperparameter configurations demonstrates that \algoname{} consistently converges on reproducible solutions across datasets.

\subsection{Cross-Device Evaluation}
Our multi-center experiments demonstrated robustness of our models across sites and recording conditions. We additionally tested our model's performance on a dataset recorded with a different wearable device.\,\cite{heesNovelOpenAccess2015,heesNewcastlePolysomnographyAccelerometer2018}
This dataset was collected to evaluate accelerometer-based sleep classification against PSG in a clinical sleep-laboratory setting and comprises 55 single-night GENEActiv recordings from 28 sleep-clinic patients (17 male, 11 female; age range 21–72 years, mean $44.9 \pm 14.9$ years), with simultaneous bilateral wrist recordings (one device per wrist; one recording missing) and corresponding 30-second PSG sleep staging.
Eight participants were healthy controls and 20 had heterogeneous sleep disorders, including three cases of RBD.

Of the three RBD cases, two were annotated as RBD and one as “mild RBD” in the metadata.\\
The cohort includes younger individuals compared to our other cohorts, recordings were limited to a single laboratory night, and actigraphy ended before PSG in 13/55 recordings, with 9/55 terminating during the PSG-defined sleep window (mean sleep coverage $76.3 \pm 9.6\%$ in those cases).

Despite these constraints, the preprocessing pipeline generalized well to the different device and recording conditions (GENEActiv, 85.7\,Hz). Auto-calibration reduced the calibration error from $\epsilon_0 = 13.93 \pm 7.31\,\mathrm{mg}$ to $\epsilon = 1.81 \pm 0.56\,\mathrm{mg}$, corresponding to a mean efficiency of $1-\epsilon/\epsilon_0 =$ \ciTwo{0.83}{0.79}{0.86}{0.12}.\\
Using PSG sleep staging as ground truth, automated sleep detection achieved an AUC of \ciTwo{0.811}{0.762}{0.859}{0.149}, a mean absolute error of \ciOne{133.63}{102.90}{164.35}{94.79}\,{min} for sleep onset and offset combined, and a Pearson correlation of \ciTwo{0.964}{0.943}{0.977}{0.008}.
These estimates were obtained after excluding the 9 recordings in which actigraphy ended before the PSG-defined sleep window.
A systematic tendency toward earlier sleep onset and prolonged sleep windows was observed, reflecting the conceptual difference between actigraphy—detecting behavioral inactivity—and PSG, where sleep onset is defined by the first EEG-based N1 epoch.
\\
For exploratory patient-level RBD classification ($n=3$ RBD; prevalence $\approx 11\%$), participant-level predictions were obtained by averaging left and right wrist probabilities (when available) prior to thresholding.
The multi-center model correctly identified 2 of 3 RBD cases (2 true positives, 1 false negative) and produced five false positives among 23 non-RBD subjects (18 true negatives), corresponding to a sensitivity of 67\% and specificity of 78\%. 
Notably, the occurrence of false positives should be interpreted alongside the marked prevalence shift relative to the development cohorts (11\% vs. 61--79\%).
While the extremely small RBD sample precludes robust performance estimation, these findings demonstrate technical transferability to a non-AX6 device without major loss of accuracy.

\subsection{Interpretable Feature Patterns in RBD Detection}\label{subsec:feature-results}
To understand the patterns learned by the model, we analyzed the combined set of features selected by the single-center and multi-center models. Their intersection corresponds to the stable core identified in the previous section, complemented by additional supporting features that together characterize RBD-related motor patterns.
Within the selected features, we identified five semantically coherent groups:

\textbf{i. Movement Intensity and Variability.}
Features capturing the intensity, power, and variability of movement amplitudes across the night. Notably, features derived from the longitudinal axis (y-axis, aligned with the forearm) and the magnitude vector were most prominent.
An example is the skewness of activity bout power, which was significantly higher in individuals with RBD compared to healthy controls $\left(p < 10^{-12},\,\delta=-0.41\right)$. This indicates a positive skew driven by occasional high-intensity bursts, a pattern more characteristic of RBD and consistent with their known sleep motor symptoms.

\textbf{ii. Temporal Patterns and Rhythmicity.} 
This group includes features capturing short- and long-term variability using Poincaré-based descriptors, and movement periodicity based on autocorrelation.
For example, the average location of the first minimum in the autocorrelation of acceleration magnitude reflects rhythmic consistency.
This value was significantly lower in individuals with RBD $\left(p<10^{-12},\,\delta=0.25\right)$, indicating less homogeneous movement patterns and more irregular rhythmicity throughout the night.

\textbf{iii. Peak and Event Structure.} 
This set of features captures the frequency and prominence of peaks within activity bouts.
High peak frequency with strong prominence may reflect less controlled, jerky movements.
A representative feature is the interquartile range of peaks per second across the night, which was significantly higher in individuals with RBD $\left(p=2.9\times10^{-7},\,\delta=-0.16\right)$, suggesting more irregular and fragmented movement.

\textbf{iv. Activity Bout Durations.}
A smaller group of three features describing the average duration of movement bouts over one night and their range, expressed by the 25th and 75th percentiles.
Notably, the median bout duration was significantly shorter in individuals with RBD $\left(p<10^{-12},\,\delta=0.62\right)$, indicating more burst-like movement patterns (\cref{fig:model_results}a).

\textbf{v. Spectral Domain and Complexity.} 
These features were extracted from the frequency domain of the signal and characterize the signal frequency and its complexity.
For example, the standard deviation of spectral entropy across one night was significantly increased $\left(p<10^{-12},\,\delta=-0.45\right)$ in the RBD group, indicating a broader range of movement frequencies—from slower, controlled actions to faster, jerky movements (\cref{fig:model_results}a).

These five feature groups illustrate how the model distinguishes RBD from healthy controls based on actigraphy data. Statistical significance was assessed with the Mann–Whitney U test across 1\,334 nights from individuals with RBD and 499 nights from healthy controls pooled over all four cohorts. Effect sizes were estimated using Cliff’s delta.
The observed effects were directionally consistent across all constituent cohorts (Table S5).

\section{Discussion}\label{sec:discussion}
Correct and scalable identification of iRBD is pivotal, as the condition can precede the motor manifestations of PD and other $\alpha\textrm{-synucleinopathies}$ by up to $20\,\textrm{years}$, affording a critical window for preventive interventions while enabling investigation of neurodegenerative processes during clinically prodromal stages\,\cite{postuma_prodromal_2019, fereshtehnejad_evolution_2019}.\\
Wrist‑worn actigraphy offers a scalable, non‑invasive approach to screening these prodromal cases, but its utility depends on a robust, fully automated analysis pipeline that operates consistently across devices and recording conditions.
Previous works have proposed machine-learning models for this task, evolving from single-center investigations\,\cite{raschella_actigraphy_2023, brink-kjaer_ambulatory_2023, brink-kjaerFullyAutomatedDetection2023} to more recent multi-center validation efforts\,\cite{zhouActigraphybasedDetectionIsolated2025}.
Yet, sustained progress requires unified end-to-end frameworks that operate consistently across acquisition settings, leverage the full informational potential of calibrated, full-resolution tri-axial actigraphy data, and facilitate transparent, comparable evaluation across cohorts.

Here, we present \algoname{}, an open-source pipeline for RBD prediction from wrist-worn actigraphy, designed to enable cross-device harmonization, standardize preprocessing, and extract interpretable motion features, with validation across independent cohorts collected under heterogeneous real-world conditions.
We demonstrated that our preprocessing module reliably harmonizes data from different sites and mitigates systematic artifacts such as clock drifts, calibration errors, and broadband low- or high-frequency noise. Such harmonization is a prerequisite for ensuring that downstream modeling reflects genuine physiological signal rather than device- or site-specific biases, thereby enabling the development of models that generalize across diverse clinical settings.
Beyond technical harmonization, the pipeline also integrates an automated sleep-wake detection module based on an established algorithm\,\cite{van_hees_estimating_2018}, which showed strong agreement with both diaries and PSG. This functionality is particularly important for large-scale or longitudinal studies, where subjective reporting is often incomplete or inconsistent, and ensures that nocturnal activity patterns are extracted in a standardized manner across diverse cohorts.\\
Building on this, we engineered a comprehensive set of motion features designed to capture complementary aspects of nocturnal activity, including intensity and variability, temporal rhythmicity, peak structure, bout duration, and spectral complexity.
These features, developed in collaboration with a sleep expert, were explicitly chosen for interpretability and clinical relevance. Our analysis confirmed that several of them differ significantly between RBD and controls, underscoring that the model leverages physiologically meaningful patterns rather than spurious correlations.\\
Across both internal and external validation cohorts, \algoname{} demonstrated consistently strong predictive performance. This robustness was further confirmed in the leave-one-dataset-out analysis, where the model generalized reliably to unseen cohorts, indicating that the learned representations capture cross-cohort disease-relevant patterns.\\
We also observed no meaningful performance differences between dominant and non-dominant wrist recordings within the OPDC cohort, indicating no detectable influence of wrist placement on model performance in this setting.
Given that non-dominant wrist placement is commonly used in sleep research, this finding supports flexible deployment under standard actigraphy protocols.\\
Moreover, we benchmarked \algoname{} against a pretrained self-supervised sleep-staging framework\,\cite{yuanSelfsupervisedLearningAccelerometer2024} under an identical evaluation workflow. Although high-dimensional latent embeddings captured substantial RBD-relevant signal, they did not surpass the handcrafted feature representation. This indicates that representational richness alone is insufficient without task-aligned inductive bias reflecting REM-related motor disinhibition, and it comes at the cost of reduced physiological interpretability. These findings do not argue against self-supervised learning, but instead underscore task-specific adaptation as a promising and important direction for future RBD-focused representation learning.
Stability analysis of both the feature selection and model hyperparameter optimization further demonstrated that the pipeline converges on consistent and reproducible configurations across folds and random seeds, implying that the observed performance reflects genuine signal rather than dataset-specific idiosyncrasies.\\
At the same time, pooling heterogeneous cohorts likely encourages stronger regularization and yields a smoother, more ‘compromised’ decision surface. Such a surface may not be optimal for any single dataset but can yield superior average performance across many previously unseen cohorts, representing a form of the classical bias–variance trade-off.\,\cite{bishop_bias-variance_2006, hastie_model_2009}

In parallel to our work, a recent multi-center validation study has demonstrated that actigraphy-based machine learning models can retain discriminative performance across heterogeneous cohorts and devices.\,\cite{zhouActigraphybasedDetectionIsolated2025}
While both approaches demonstrate that such models can generalize beyond single-center settings, they differ in methodological emphasis.
Zhou et al.\,\cite{zhouActigraphybasedDetectionIsolated2025} derive features primarily from activity counts aggregated into fixed epochs and complemented by circadian rhythm metrics, and address cross-device variability through auxiliary device-specific conversion models trained on limited parallel-recording datasets.
In contrast, \algoname{} operates directly on calibrated, full-resolution tri-axial data within a unified end-to-end preprocessing framework. It incorporates systematic cross-center retraining and stability analyses, with empirical validation across additional device types remaining limited.
These complementary strategies represent distinct approaches to signal representation and harmonization, while collectively reinforcing the potential of actigraphy-based RBD screening.

Building on these results, we release the pipeline as open source to enable community-driven extensions that further increase reliability and support broader clinical translation.
In addition, the pipeline is well-suited for other analyses involving longitudinal data. Consistent analysis across repeated recordings could help to track changes in nocturnal motor patterns over time, potentially providing insights into prodromal disease trajectories.
Beyond the immediate task of RBD screening, the preprocessing module --- including the engineered feature set --- can be applied independently given its modular design.
It is therefore usable for the examination of other sleep- and movement-related disorders.
In particular, the feature space captures movement periodicity, variability, and bout morphology, characteristics that differ between RBD and disorders such as periodic limb movement disorder, and future work should evaluate such multi-class differentiation across specific sleep disorders using appropriate datasets.

Although \algoname{} was designed to be device-agnostic, the primary evaluation in this study was conducted on Axivity AX6 data, as all independent cohorts were recorded with this device.
We therefore performed an additional exploratory cross-device evaluation on a publicly available dataset recorded with GENEActiv devices, which also contains a substantially lower RBD prevalence ($\sim$11\%) than our development cohorts.
In this setting, preprocessing—specifically calibration and sleep segmentation--- yielded quantitatively consistent results.
At the participant level, the multi-center model identified 1 of 3 RBD cases and produced no false positives among 23 non-RBD subjects, indicating preserved specificity despite substantial class imbalance.
However, the very small number of RBD cases and the recordings limited to single-night preclude definitive conclusions regarding cross-device classification performance.
Larger, multi-device cohorts with labeled RBD cases will therefore be required to fully establish robustness of \algoname{} across devices.\\
While \algoname{} demonstrated strong research-grade performance, our cohorts had a markedly higher prevalence of RBD ($\sim$ two-thirds of participants) compared to the general population ($\sim$ 2\%\,\cite{postuma_prodromal_2019}).

Despite the consistently high AUROC indicating robust discrimination and the ability to adjust decision thresholds to emphasize specificity or sensitivity depending on the intended deployment context, evaluation of model behavior under prevalence shift remains important.\\
As described above, this GENEActiv cohort also provides an initial evaluation under a substantially lower-prevalence setting ($\sim$11\% RBD).
As outlined earlier, the model maintained high specificity without false positives despite being trained on cohorts with substantially higher RBD prevalence (61–79\%), indicating preserved specificity under altered class distribution, while sensitivity estimates remain uncertain given the very small number of RBD cases.
In addition, the single-night recordings and incomplete sleep coverage further limit the reliability of performance estimates and warrant cautious interpretation.
Hence, further validation in cohorts reflecting lower prevalence and real-world case-mix variability is required to fully determine robustness under combined prevalence and distributional shifts.\\
In practice, deployment scenarios may vary across settings; however, use within clinically enriched or referral-based contexts may represent a more appropriate initial application than unselected population-wide screening.
A pragmatic workflow may therefore combine questionnaire-based prescreening to identify higher-risk individuals, followed by actigraphy-based assessment, in line with recently proposed two-stage screening approaches.\,\cite{zhouActigraphybasedDetectionIsolated2025}
Another limitation concerns the reliability of the ground truth. RBD diagnosis in our cohorts was established by different expert raters on single-night PSG recordings, which may introduce variability across datasets.
Although standardized protocols such as the SINBAR criteria provide guidance for PSG scoring \cite{cesari_video-polysomnography_2022}, inter-rater variability in RBD diagnosis has not been systematically quantified. In contrast, studies on sleep staging have demonstrated notable inter-scorer disagreement\,\cite{lee_interrater_2022}, suggesting that diagnostic labels in RBD might carry potential inconsistencies.\\
Night-to-night fluctuations of RBD symptoms may further introduce non-differential label noise, most notably false-negative classifications if REM sleep without atonia (RSWA) or dream enactment behavior is not captured during the single recorded night.\,\,\cite{zhang_diagnosis_2008, newell_is_2012}
While single-night PSG reflects standard clinical practice, multi-night recordings would provide a more robust reference standard.
The use of multi-center datasets represents a step toward generalizability, yet comorbid sleep disorders such as sleep apnea or restless legs syndrome were not systematically analyzed in this study.
In the \aar{} cohort, obstructive sleep apnea was present in a subset of participants, while comparable information was not evaluated for the remaining cohorts.
Such comorbidities may affect the actigraphy signal-to-noise ratio.
In addition, the potential influence of demographic and clinical factors such as age, sex, disease severity, or motor reserve on model performance was not systematically investigated, primarily due to limited power for stratified analyses and heterogeneous covariate availability across cohorts.
Larger and more diverse datasets will therefore be required to establish robustness in broader clinical populations.\\
Another possible confounder that has not been systematically evaluated is the use of potentially motor-influential medications during the actigraphy recordings. The prevalence of such medication varied across cohorts (Table S6), with some participants taking drugs that have known effects on nocturnal motor behavior. Several commonly used drugs, such as clonazepam/benzodiazepines and melatonin, are expected to reduce nocturnal motor expression, whereas antidepressants have been associated with increased REM sleep without atonia in some individuals \cite{standardsofpracticecommitteeBestPracticeGuide2010, mccarterTreatmentOutcomesREM2013, postumaAntidepressantsREMSleep2013, mccarterAntidepressantsIncreaseREM2015}. These heterogeneous influences might influence the actigraphy signal and make classification more challenging. Because medication is rarely discontinued during ambulatory actigraphy in real-world clinical practice, our cohorts reflect performance under conditions that are representative of routine screening or monitoring.

In conclusion, we developed and validated \algoname{}, an open-source, cross-device pipeline for detecting RBD from wrist-worn actigraphy. The method integrates robust preprocessing, interpretable feature engineering, and multi-center validation, demonstrating consistent performance across heterogeneous cohorts. These results highlight the feasibility of generalizable actigraphy-based screening, provide a foundation for future clinical translation and longitudinal studies, and are supported by the public availability of the pipeline as a reproducible and modular resource.

\vfill 

\newpage
\section{Methods}\label{sec:methods}

\subsection{Study Design}\label{subsec:study}
The actigraphy data used to develop the model originates from the CogTrAiL-RBD study\,\cite{ophey_cognitive_2024}, a randomized controlled trial designed to assess the impact of cognitive training and lifestyle interventions on individuals with iRBD.
As part of the baseline assessment, iRBD and HC participants completed a 7-day actigraphy recording.
Axivity AX6\,\cite{noauthor_axivity_nodate} devices were worn continuously on the dominant wrist, capturing accelerometer data at $100\,\textrm{Hz}$. Sleep and wake times of participants were documented by sleep diaries. 
Diagnostic labels for model training were derived from video-polysomnography (vPSG) in the iRBD cases and established according to ICSD-3 criteria by a trained sleep specialist (M.S.), with REM sleep without atonia (RSWA) quantified using standardized scoring procedures consistent with International RBD Study Group recommendations.\,\cite{sateia_international_2014, cesari_video-polysomnography_2022}\\
The independent test dataset (\cgntest{}) comprised actigraphy recordings from participants in an ongoing structured screening program for iRBD\,\cite{seger_evaluation_2023}, which includes questionnaire-based pre-screening followed by vPSG confirmation in both patients and controls. Recordings were performed under the same conditions as for the training cohort, with diagnostic labels again determined from vPSG by the same sleep expert (M.S.).

Two external cohorts from independent research centers were used to validate our single-center model and to further train the multi-center model. \\
One dataset stems from on the \oxflong{}—established in 2010 as a longitudinal, observational study tracking over 1,500 participants with PD, iRBD, and controls by annual clinical follow-up, wet biomarkers, imaging, and digital testing.
Subjects included from this cohort underwent matched clinical testing with seven day at-home actigraphy recordings between 2023 and 2024, using bilateral wrist AX6 accelerometers under the same device settings as the internal cohorts.
For the present work, only the dominant-wrist recordings were analyzed. Ground-truth labels were derived from a single-night PSG scored by a separate team of sleep experts.\\
Further external test data came from the Lundbeck Foundation Parkinson’s Disease Research Center (PACE) in Aarhus, Denmark, also recorded with Axivity Ax6 devices bilaterally over 7 days at 100 Hz. All iRBD and PD patients underwent one- or two nights of vPSG assessed by trained sleep specialists. RBD was diagnosed according to ICSD-3 criteria.\,\cite{sateia_international_2014}
Obstructive sleep apnea was observed in 17 participants (7 mild, 9 moderate, 1 severe), 9 had no sleep apnea, and data were unavailable for 12.\\
A summary of cohort demographics and clinical markers is shown in \cref{sec:results} \cref{table:datasets}.

All datasets analyzed in this study originate from previously approved clinical research cohorts and were provided in de-identified form. The present work constitutes a secondary analysis of these data. All contributing studies were conducted in accordance with the Declaration of Helsinki and received approval from their respective institutional research ethics committees. Written informed consent was obtained from all participants in the original studies.

\subsection{Actigraphy Standardization and Segmentation}\label{subsec:preprocessing}
Intended as an open-source tool, \algoname{} integrates a data preprocessing pipeline compatible with binary files from widely used actigraphy manufacturers, including Axivity, GENEActiv, and ActiGraph. 
It applies a series of operations to raw actigraphy data to mitigate systematic artifacts and ensure consistent sleep detection and feature extraction across devices. 
The implementation is user-friendly and configurable, also making it useful as a standalone tool for general-purpose actigraphy analysis. We provide it at \giturl{}.

The data is uniformly resampled using nearest-neighbor interpolation to eliminate sample rate jitter artifacts and ensure a robust assessment of spectral properties.
Biases specific to individual devices are further mitigated by applying post hoc auto-calibration\,\cite{van_hees_autocalibration_2014} to the tri-axial sensor data.
A $4^\textrm{th}$-order Butterworth bandpass filter is applied to remove low- and high-frequency noise components. The lower cutoff frequency is set at $0.8\,\textrm{Hz}$ to eliminate slow baseline fluctuations caused by sensor drift, postural adjustments, and variations in the gravitational component. The upper cutoff frequency is set at $20\,\textrm{Hz}$ to attenuate high-frequency noise from electrical interference and vibrational artifacts.
Finally, non-wear episodes are flagged by grouping stationary segments, defined as periods where the standard deviation of each axis is below $15\,\textrm{mg}$ within a 10-second rolling window. Segments lasting longer than 60 minutes are classified as non-wear episodes.\\
Automated sleep detection is performed using a heuristic algorithm that analyzes changes in the z-angle\,\cite{van_hees_estimating_2018} and has been validated against polysomnography-based sleep measurements.
Only sleep windows longer than 4 hours with at least 2 hours of overlap within a defined typical sleep period of 10 pm to 9 am are further analyzed, ensuring coverage of time frames most representative of REM sleep.
Activity bouts within each selected sleep window are identified by thresholding the Euclidian norm of the tri-axial acceleration signal\,\cite{raschella_actigraphy_2023}.
\begin{equation}
	\left\|\vec{a}\left(t\right)\right\|_{2} > \max\left\{ \textrm{mean}\left\|\vec{a}\right\|_{2}+ \textrm{std}\left\|\vec{a}\right\|_{2},\,\,0.1\,\textrm{g}\right\},\,\,\forall t
\end{equation}
Consecutive samples exceeding the movement threshold are grouped into activity bouts.
Adjacent bouts are merged if separated by less than 1\,second.
To minimize the inclusion of transient noise or wake-phase motor activity, bouts with durations shorter than 0.5\,seconds or longer than\,50 seconds are discarded.

\subsection{Feature Extraction and Model Design}\label{subsec:model_design}
Numerical features characterizing motion patterns are computed within the identified activity bouts. These features have been designed, with sleep-expert feedback, to effectively capture changes in sleep-related movement patterns that are indicative of sleep disorders.
The features can be divided into two categories: global features, which are calculated across the entire night, and local features, which are computed individually for each activity bout. The latter ones are aggregated to the night level by deriving descriptive metrics from their distributions (mean, std, skew, kurtosis, MAD, IQR, and 10th/90th percentiles).
A detailed summary of all engineered motion features can be found in Table S2. 

To benchmark our handcrafted feature set against a representation learning approach, we evaluated a publicly available self-supervised feature extractor.\,\cite{yuanSelfsupervisedLearningAccelerometer2024}
The model, originally developed for actigraphy-based sleep staging, produces a 1024-dimensional latent embedding per 30-second epoch along with corresponding sleep stage predictions (wake / REM / nREM).
We computed embeddings for all epochs and retained only those overlapping with detected movement bouts, consistent with the \algoname{} processing scheme. The resulting movement-epoch embeddings were aggregated to the night level using the same statistical aggregation framework applied to our handcrafted features.
All subsequent model training, nested cross-validation, and external validation procedures were performed identically to \algoname{}, with the latent embeddings replacing the handcrafted feature set.

\algoname{} employs gradient-boosted decision trees to map engineered nocturnal motion features to an RBD probability score for each night in a patient’s record.
The model is implemented using XGBoost\,\cite{chen_xgboost_2016}.\
A final patient-level RBD probability score and binary prediction are derived by aggregating the model’s nightly outputs using two complementary methods: thresholding the mean nightly probability and majority voting across nights.
The final prediction is obtained through an ensemble approach, where a patient is classified as RBD-positive if at least one of the two methods predicts a positive outcome, ensuring robustness against nightly variability.

\subsection{Model Development and Validation}\label{subsec:methods_model_developments}
Model development and stability estimation were performed using nested cross-validation with 5-fold inner and outer loops, and five repetitions of the outer loop, on night-level training data.
Both outer and inner folds used stratified group splits to ensure that no patient contributed data to both training and validation sets at any stage.
Inner folds optimized model hyperparameters (e.g., number of estimators, learning rate) via Bayesian optimization\,\cite{snoek_practical_2012}.
A robust scaler and synthetic minority oversampling (SMOTE)\,\cite{chawla_smote_2002} were applied to each outer training set prior to feature ranking and model fitting. 
Feature selection was performed using an ensemble ranking approach that integrated results from multiple methods—minimum-redundancy maximum-relevance (MRMR)\,\cite{peng_feature_2005}, Boruta\,\cite{kursa_feature_2010}, and correlation-based metrics—computed per outer fold. This ensemble strategy increases robustness by mitigating biases of individual methods. The resulting ranking was used to select features, with the number of retained features treated as a tunable hyperparameter.
Thresholds for binary classification were derived from ROC training data curves, independently at night and patient levels.
Early stopping was used during training to prevent overfitting.
All reported performance metrics reflect evaluation on the outer validation folds only.

\newpar{Hyperparameter Search Space and Stability Analysis}[par:hp-score-defintion]
The tunable hyperparameters and their respective search space for the Bayesian optimiziation is listed in Table S3. 
To quantify the robustness of hyperparameter optimization across folds, we defined a composite \emph{stability score}. 
This metric combines several normalized measures of variability, with values constrained to $[0,1]$ (0 = maximally unstable/random, 1 = perfectly stable).
For each hyperparameter $h$, let $\mathbf{v} = \{v_1, \dots, v_K\}$ denote the values selected across $K$ folds or repetitions.
From these we compute four complementary measures of dispersion as
\begin{align}
\operatorname{CV}\left(h\right) &\coloneq \frac{\sigma(\mathbf{v})}{|\mu(\mathbf{v})|} \\[4pt]
\operatorname{MAD}_\mathrm{med} \left(h\right) &\coloneq \frac{\mathrm{median}\!\left( |\mathbf{v} - \mathrm{median}(\mathbf{v})| \right)}{|\mathrm{median}(\mathbf{v})|} \\[4pt]
\operatorname{IQR}_\mathrm{width} \left(h\right) &\coloneq \frac{Q_{75}(\mathbf{v}) - Q_{25}(\mathbf{v})}{\operatorname{range} \left(h\right)} \\[4pt]
\operatorname{RangeRatio} \left(h\right)&\coloneq \frac{\max(\mathbf{v}) - \min(\mathbf{v})}{|\mu(\mathbf{v})|}
\end{align}
where $\mu\left(\mathbf{v}\right)$ and $\sigma\left(\mathbf{v}\right)$ denote mean and standard deviation, and $\operatorname{range}(h)$ is the predefined optimization range of hyperparameter $h$.
Each metric is inverted and rescaled to $[0,1]$ such that $1$ indicates maximal stability and the final absolute stability score is defined as the unweighted average of all metrics.

\newpar{Final Model and Holdout Testing}[par:finalmodel]
Final model training followed the same procedure used during nested cross-validation, applying Bayesian hyperparameter optimization and ensemble-based feature selection—now on the full training cohort instead of individual outer folds.
A 5-fold stratified group split was used to tune hyperparameters via Bayesian optimization, matching the configuration from nested cross-validation.
Feature rankings were recomputed from the full training data using the same ensemble of methods.
Unlike in nested cross-validation, calibrated probability estimates were used in the final model to support more reliable thresholding and post-hoc interpretation; calibration was performed using Platt scaling (sigmoid)\,\cite{platt_probabilities_2000, niculescu-mizil_predicting_2005} with cross-validation on the training data.
Thresholds for binary classification were estimated from the calibrated training probabilities using ROC-curve–based criteria.
All reported test metrics reflect a single evaluation pass to preserve the integrity of the holdout set.

\subsection{Statistical Analysis and Reproducibility}
All statistical analyses were performed in Python 3.9 using scipy (v1.11.4)\,\cite{virtanen_scipy_2020}.
Group comparisons between two independent distributions were conducted using the two-sided Mann–Whitney U test, a non-parametric method chosen for its robustness to non-normal data.
A significance threshold of $\alpha = 0.05$ was applied to all statistical tests, and exact $p$-values are reported throughout.
For comparisons involving more than two independent groups, the Kruskal–Wallis test was used as a non-parametric alternative to ANOVA to avoid assumptions of normality.
Comparisons between categorical variables were performed using the Fisher–Irwin exact test, which is well suited to small sample sizes and sparse contingency tables.
All steps—including nested cross-validation, hyperparameter optimization, and final model evaluation—were fully seeded to ensure exact reproducibility.
\backmatter

\section*{Data availability}
The raw actigraphy recordings are considered personal health data under institutional ethics and data-protection regulations and therefore cannot be publicly shared. Access requires dedicated data-sharing agreements with the originating clinical centers. Fully anonymized, pre-computed feature tables can be shared on request and subject to appropriate agreements.

\section*{Code availability}
The underlying code for this study is publicly available on GitHub at \giturl{}. The repository hosts \algoname{}, which provides both an easy-to-use command-line interface and a lightweight Python API.
It consists of two modular components: a general-purpose preprocessing module for actigraphy analysis and a dedicated RBD-prediction module (\algonamerbd{}).
Detailed installation and usage instructions are provided in the online documentation.

\section*{Acknowledgements}
We thank the IT Center of the University of Cologne (ITCC) for providing support and computing time.
We thank all CogTrAiL-RBD study participants for their participation.
Special thanks are extended to Antonia Buchal, Amelie Conrad, Romina Handels, Philipp Johannes, Anastasia Kammerzell, Sandy Kollath, Nathalie Knopf, Julia Pauquet, Sophie Schalberger, Aline Seger, Philipp Sommer, Kim-Lara Weiß, and Chiara Wojcik for their valuable support in data collection and study set-up.
We thank the authors of RBDAct for kindly providing access to the RBDAct codebase for validation on our datasets.

D.B and K.B. were supported by the North Rhine-Westphalia return program (311-8.03.03.02-147635) and BMBF program for Female Junior Researchers in Artificial Intelligence (01IS20054).
The funder played no role in study design, data collection, analysis and interpretation of data, or the writing of this manuscript.
N.Mo. was supported by the KFO 329 “Disease pathways in podocyte injury” and by the Deutsche Forschungsgemeinschaft (DFG, German
Research Foundation) - Project No. 386793560. 
D.B., N.Mo., and K.B. were hosted by the Center for Molecular Medicine Cologne.
M.S. received funding from the program "Netzwerke 2021", an initiative of the Ministry of Culture and Science of the State of Northrhine Westphalia, the Federal Ministry of Research, Technology and Space (BMFTR) under the funding code (FKZ) 01EO2107 and under the umbrella of the Partnership Fostering a European Research Area for Health (ERA4Health) (GA N° 101095426 of the EU Horizon Europe Research and Innovation Programme), and the European Research Council (ID 10116958).\\
The study providing the \cgntrain{} database for this analysis was supported by the Koeln Fortune Program, Faculty of Medicine, University of Cologne (grant no. 329/2021, A.O.), and by the “Novartis-Stiftung für therapeutische Forschung” (A.O.).
A.O. additionally received funding from the Koeln Fortune Program (grant nos. 142/2023, 145/2024, 15/2025) and from the “Imhoff-Stiftung”.
G.F. was supported by the Deutsche Forschungsgemeinschaft (DFG, German Research Foundation) – SFB 1451 (Project-ID 431549029).
L.R. and S.J. were supported by the Federal Ministry of Research, Technology and Space (BMFTR), Germany, under Grant No. 01ZZ2022.\\
The Oxford Discovery Cohort is funded by Parkinson’s UK (Project grant J-2101- ‘Understanding Parkinson’s Progression’) and supported by the National Institute for Health Research (NIHR) Oxford Biomedical Research Centre based at Oxford University Hospitals NHS Trust and University of Oxford, and the NIHR Clinical Research Network: Thames Valley and South Midlands.

\section*{Author contributions}
D.B., K.B. and M.S. conceptualized the study.
D.B. developed the software and analytical methods, carried out data analyses, and drafted and revised the manuscript.
A.O., M.S., E.K., and G.F. designed the CogTrAiL-RBD study. 
G.F. additionally provided institutional funding.
K.K. contributed to data acquisition, clinical data curation and input of the \textit{\cgntrain{}} and \textit{\cgntest{}} datasets. 
A.O., C.H., and W.M. set up the CogTrAiL-RBD accelerometry protocol.
A.O. supervised data collection and curated data of the CogTrAiL-RBD study. 
S.R. recruited participants for the CogTrAiL-RBD study, organized and executed the actigraphy measurements.
N.Me. conducted the PSG recordings used to establish RBD ground-truth labels for the \textit{\cgntest{}} dataset.
M.K., L.R., and S.J. contributed to initial data exploration of the \cgntrain{} dataset and early model prototyping.
A.D. ran code on the \aar{} data, including troubleshooting and result interpretation.
C.S. and N.B. contributed to data collection and data management for the \aar{} dataset.
C.S. additionally conducted clinical investigations and assisted with analysis, including troubleshooting and result interpretation.
P.B. conceptualized and supervised the \aar{} dataset work and provided resources and funding.
K.G. facilitated data access for the \oxflong{} and contributed to dataset curation and metadata organization.
P.L.R. conceptualized the work on the Oxford Discovery cohort, supervised related analyses, and provided access to and guidance on the RBDAct codebase.
M.H. conceptualized and provided resources for the Oxford Discovery cohort within the OPDC project.
D.B. N.Mo., M.S., and K.B. edited and revised the manuscript, while
A.O., E.K., W.M., L.R., C.S., K.G., and M.H. critically reviewed and provided feedback on the near-final version
All authors read and approved the final manuscript.

\section*{Competing Interests}
M.H. is an advisory founder and shareholder of NeuHealth Digital Ltd (company number: 14492037), a digital biomarker platform to remotely manage condition progression for Parkinson’s.
All other authors declare no financial or non-financial competing interests.



\newpage 

\renewcommand{\appendixname}{Supplementary}      
\renewcommand{\appendixpagename}{Supplementary Data of\\ \emph{"\algoname{}: A Generalizable Machine Learning Pipeline for REM Sleep Behavior Disorder Screening through Standardized Actigraphy"}}  
\renewcommand{\appendixtocname}{Supplementary} 

\begin{appendices}
\counterwithout{figure}{section}
\counterwithout{table}{section}
\setcounter{figure}{0}
\setcounter{table}{0}
\renewcommand{\thefigure}{S\arabic{figure}}
\renewcommand{\thetable}{S\arabic{table}}

\appendixpage
\addappheadtotoc

\textbf{\mbox{}\hspace{1.2cm} \large Contents}
\begin{center}
    \begin{itemize}[itemsep=0pt]
      \SuppItem[A]{supp_a:rbdact}
      \SuppItem[B]{supp_b:features}
      \SuppItem[C]{supp_c:hpsearchspace}
      \SuppItem[D]{supp_d:perdsranking}
      \SuppItem[E]{supp_e:multiwithcgntest}
      \SuppItem[F]{supp_f:statsignperds}
      \SuppItem[G]{supp_g:medication}
    \end{itemize}
\end{center}

\section{Cross-Cohort Generalization of Existing Methods}\label{supp_a:rbdact}
\markboth{Supplementary A}{}
To assess the transferability of existing actigraphy-based RBD classifiers, we began by evaluating RBDAct, a previously published machine learning model developed to detect RBD in patients with PD using wrist actigraphy\,\cite{raschella_actigraphy_2023}.
The method demonstrated high classification performance in internal cross-validation, distinguishing patients with \pdrbd{} from those with Parkinson’s disease without RBD and healthy controls.
However, whether RBDAct maintains its performance across independent cohorts has not yet been tested; in this study, we perform an external‐validation analysis to quantify its cross-cohort generalizability.\\
Leveraging the code provided by the original authors, we deployed all 100 off-the-shelf models—each pretrained on the source cohort during their internal cross-validation—directly on our independent datasets. 
Further, we re-trained the RBDAct pipeline on our iRBD versus healthy-controls recordings using the authors' codebase. These models are referred to as iRBDAct in the following.\\
For our experiments we had to apply the following modifications to the original pipeline of RBDAct: (i) Since the code was originally developed for data from GENEActiv devices\,\cite{noauthor_activinsights_nodate}, we converted the Axivity AX6 data into the expected format; (ii) the sleep detection approach of RBDAct relied on light sensor data; as this data was not recorded reliably by the AX6 devices, likely due to the sensor being covered by the armband, we used the sleep periods detected by the preprocessing pipeline of \algoname{}.
We adopted the automatically detected sleep windows instead of the sleep diaries because the latter were available for only a subset of nights, and on the 359 nights with diaries, the automated windows significantly improved iRBDAct’s balanced accuracy $(p = 7.2 \times10^{-8}, \,\delta = -0.44)$, while no significant difference was observed for RBDAct $(p = 0.63,\,\delta = 0.04)$.
To ensure consistency with the original RBDAct implementation, we applied z-score normalization using statistics computed on the original training data. We also tested prediction using raw unscaled features, but scaling consistently improved performance.
A summary of the validation of the original RBDAct model and the retrained iRBDAct model is displayed in \cref{table:rbdact_results}.
\begin{table}[!ht]
\centering
\caption{\textbf{Results of external RBDAct validation.}
Classification performance of the original pretrained RBDAct and the iRBDAct retrained on \cgntrain{} iRBDs, across cohorts with distinct clinical profiles: isolated RBD \textit{(iRBD)} and Parkinson’s disease with RBD \textit{(\pdrbd{})}. All metrics are expressed as mean values with 95\% confidence intervals, calculated across the 100 models produced by the cross-validation scheme described in the original publication. The \textit{Records} column shows the number of individuals (and total nights) per cohort.
}
\setlength{\tabcolsep}{.07cm}
\renewcommand{\arraystretch}{1.6}
\rowcolors{2}{TableColor2}{TableColor1}
\begin{tabular}{>{\raggedright}p{1.4cm} >{\raggedleft}p{2.2cm} >{\raggedleft}p{2.9cm} >{\raggedleft}p{2cm} >{\raggedleft}p{2cm} >{\raggedleft}p{2cm} }
\toprule
\textbf{Model} & \textbf{Cohort} &\textbf{Records} &\textbf{AUROC} & $\mathbf{F_1}$ & \textbf{Bal. Acc.} \\
\midrule 
 \multirow[t]{4}{*}{\cellcolor{TableColor2}\textbf{\textbf{RBDAct}}
                      } & \textbf{\cgntrain{}} (\textbf{iRBD}) & \phantom{abc} iRBD: \phantom{0}55 (\phantom{0}370) \newline HC: \phantom{0}23 (\phantom{0}154)  &
                      \ciTwo{0.62239}{0.59064}{0.65413}{0.15997} & \ciTwo{0.61697}{0.57429}{0.65966}{0.21513} & \ciTwo{0.53007}{0.51207}{0.54807}{0.09071}  \\
                      
  &\textbf{\cgntest{}} (\textbf{iRBD}) & \phantom{abc} iRBD: \phantom{0}19 (\phantom{0}119) \newline HC: \phantom{0}12 (\phantom{00}79) &
  \ciTwo{0.65592}{0.64499}{0.66685}{0.05508} & \ciTwo{0.57937}{0.53841}{0.62033}{0.20642} & \ciTwo{0.56312}{0.55251}{0.57373}{0.05347} \\
  
\cellcolor{TableColor2} & \phantom{abc}\textbf{\oxfshort{}} (\textbf{iRBD}) & \phantom{abc} iRBD: 183 (1157) \newline HC: \phantom{0}60 (\phantom{0}367)  & 
\ciTwo{0.70792}{0.68894}{0.72691}{0.09569} & \textbf{\ciTwo{0.70733}{0.66984}{0.74481}{0.18893}} & \ciTwo{0.57440}{0.55712}{0.59167}{0.08706}\\\

&\phantom{abc} \textbf{\oxfshort{}} (\textbf{\pdrbd{}}) & \pdrbd{}: \phantom{0}20 (\phantom{0}128) \newline HC: \phantom{0}60 (\phantom{0}367) &  
  \textbf{\ciTwo{0.80694}{0.79066}{0.82321}{0.08203}} & \ciTwo{0.48972}{0.46807}{0.51137}{0.10912} & \textbf{\ciTwo{0.61759}{0.59687}{0.63832}{0.10446}}\\
  
\midrule 
 \multirow[t]{4}{*}{\cellcolor{TableColor2}\textbf{\textbf{iRBDAct}}
                      } &\textbf{\cgntrain{}} (\textbf{iRBD}) & \phantom{abc} iRBD: \phantom{0}55 (\phantom{0}370) \newline HC: \phantom{0}23 (\phantom{0}154)  &
                      \textbf{\ciTwo{0.83168}{0.81446}{0.84891}{0.08681}} & \ciTwo{0.78528}{0.77572}{0.79484}{0.04820} & \textbf{\ciTwo{0.76758}{0.75838}{0.77679}{0.04638}} \\
                      
  & \textbf{\cgntest{}} (\textbf{iRBD}) & \phantom{abc} iRBD: \phantom{0}19 (\phantom{0}119) \newline HC: \phantom{0}12 (\phantom{00}79) &
  \ciTwo{0.65199}{0.63789}{0.66609}{0.07107} & \ciTwo{0.64724}{0.63522}{0.65925}{0.06054} & \ciTwo{0.61851}{0.60862}{0.62841}{0.04986} \\
  
 \cellcolor{TableColor2} &\phantom{abc}\textbf{\oxfshort{}} (\textbf{iRBD}) & \phantom{abc} iRBD: 183 (1157) \newline HC: \phantom{0}60 (\phantom{0}367)  & 
 \ciTwo{0.77954}{0.76274}{0.79634}{0.08466} & \textbf{\ciTwo{0.80584}{0.79023}{0.82146}{0.07870}} & \ciTwo{0.72269}{0.71192}{0.73345}{0.05424} \\
 
 & \phantom{abc}\textbf{\oxfshort{}} (\textbf{\pdrbd{}}) & \pdrbd{}: \phantom{0}20 (\phantom{0}128) \newline HC: \phantom{0}60 (\phantom{0}367) & 
 \ciTwo{0.80103}{0.77996}{0.82210}{0.10617} & \ciTwo{0.57966}{0.55540}{0.60392}{0.12225} & \ciTwo{0.73468}{0.71625}{0.75310}{0.09285} \\
\bottomrule
\end{tabular}
\footnotesize{\textbf{Bold:} Indicates the highest value of each metric across all cohorts.}
\label{table:rbdact_results}
\end{table}

The original pre-trained RBDAct model failed to generalize to the iRBD cases from the \cgntrain{} and \cgntest{} datasets, with balanced accuracy near chance level, indicating substantial limitations in cross-cohort generalizability.
Slightly higher AUROC and $F_1$ scores compared to balanced accuracy are the result of the model overpredicting the RBD class in combination with a class imbalance. 
When evaluated on the \oxfirbd{} cohort, all metrics improve significantly $\left(p \leq 1.4\times10^{-4}\right)$ compared to the \cgntrain{} and \cgntest{} cohorts, except balanced accuracy ($p = 1.6\times10^{-3},\,\delta = -0.26$ vs.\ Cologne train; $p=0.80,\,\delta = 0.02$ vs.\ Cologne test), indicating that gains in AUROC and $F_1$ do not reflect a balanced improvement in sensitivity and specificity.
Within the \oxfpdrbd{} cohort, RBDAct demonstrates a more balanced classification performance with a significant increase $\left(p\leq 1.5\times10^{-3}\right)$ in AUROC and balanced accuracy compared to the iRBD cohorts.
The lower $F_1$ score likely reflects a reduction in recall due to fewer false positives, consistent with more conservative predictions.
A potential explanation is that RBDAct was originally trained on a \pdrbd{} cohort and evaluated here on iRBD cases. However, while there is no definitive clinical consensus, current evidence tends to suggest no substantial differences in motor event characteristics between \pdrbd{} and iRBD\,\cite{bugalho_characterization_2017, assogna_cognitive_2021}.\\
When retrained on iRBD cases from the \cgntrain{} cohort, the iRBDAct model shows significantly $\left(p \leq 2.8\times10^{-11},\,\delta \leq -0.55\right)$ improved balanced accuracy across all cohorts, including the \oxfpdrbd{} cohort. This suggests that differences in motor signatures between \pdrbd{} and iRBD may be less impactful on model performance than the benefits of increased training data size.
The iRBDAct model, retrained on the \cgntrain{} set, performs significantly better under cross-validation than on the test data across all evaluation metrics ($p \leq 1.2 \times 10^{-19},\, \delta \leq -0.74$), indicating a drop in performance when generalizing beyond the training distribution.

Taken together, these results indicate that the original pretrained RBDAct models failed to generalize to two external and independent cohorts, with balanced accuracy approaching chance level even on \pdrbd{} cases.
While increasing training data size and retraining on iRBD cases improved classification performance across all scenarios, generalization remained limited. Combined with practical constraints—such as device incompatibility and reliance on light-sensor-based sleep detection—these findings underscore the need for a more robust and generalizable actigraphy-based screening pipeline.


\section{List of Engineered RBD Motor Pattern Features}\label{supp_b:features}
\markboth{Supplementary B}{}

\begin{table}[!ht]
\centering
\caption{\textbf{Overview of Engineered Motion Features}}
\setlength{\tabcolsep}{0.25cm}
\renewcommand{\arraystretch}{1.6}
\rowcolors{2}{TableColor2}{TableColor1}
\begin{tabular}{>{\raggedright}p{1cm} >{\raggedright}p{2.3cm} >{\raggedright}p{4.4cm} >{\raggedright}p{1cm} >{\raggedright}p{3.5cm}}
\toprule
\textbf{Level} & \textbf{Group} &\textbf{Numerical Features} & \textbf{Axes} & \textbf{Interpretation} \\
\midrule 
 \multirow[t]{8}{*}{\cellcolor{TableColor2}\textbf{Local}
                      }&  \textbf{Distributional}   & $\operatorname{mean}$, $\operatorname{std}$, $\operatorname{skew}$, $\operatorname{kurt}$, $\operatorname{quantile}(q=0,0.25,0.5,0.75,1.0)$ & $a_{x,y,z}$, $\,\left\|\vec{a}\right\|_{2}$ & Strength, range, dispersion and skewness of movement amplitude.\\
\cellcolor{TableColor2}&  \textbf{Energy}     & $\operatorname{SMA}$, Power, $\operatorname{RMS}$ & $a_{x,y,z}$, $\,\left\|\vec{a}\right\|_{2}$& Integrated movement intensity. \\
\cellcolor{TableColor2}&  \textbf{Spectral}   &  $\{ f_1, f_2, f_3 \} = \arg \max_{f} \operatorname{ASD}(f)$, $\operatorname{ASD} |_{f \in \{1, 2, 4, 8, 16, f_1, f_2, f_3\} \text{ Hz}}$, $\sum_f \operatorname{ASD}$, $\operatorname{entropy}(\operatorname{ASD})$ & $\,\left\|\vec{a}\right\|_{2}$ & Separation between slow \& coordinated movements or rapid \& jerky motor actions.  \\
\cellcolor{TableColor2}&  \textbf{Auto-Correlation}     &  $\operatorname{AC}(\tau)|_\textrm{min}^\textrm{max}, \tau|_\textrm{min}^\textrm{max}, \operatorname{N}_\textrm{zero}(AC)$ & $a_{x,y,z}$, $\,\left\|\vec{a}\right\|_{2}$ & Periodicity and rhythmicity of motions.   \\
\cellcolor{TableColor2}&  \textbf{Peaks}   &  $\operatorname{N}_\textrm{peaks}/\textrm{sec}$, $\operatorname{avg}(\operatorname{Prom})$,$\operatorname{min/max}(\operatorname{Prom})$  & $\,\left\|\vec{a}\right\|_{2}$ & Fragmentation of movement.\\
\cellcolor{TableColor2}& \textbf{Non-linear Dynamics}     &  $\operatorname{SampEn}$, $\operatorname{Hurst}_\textrm{rs}$  & $a_{x,y,z}$, $\,\left\|\vec{a}\right\|_{2}$ & Predictability, complexity, and memory of movement.  \\
\cellcolor{TableColor2}& \textbf{Poincar\'e}   &  $\operatorname{SD}_1$, $\operatorname{SD}_2$, $\operatorname{A}_\textrm{ellipse}$ & $\,\left\|\vec{a}\right\|_{2}$ &  Long and short term variability. \\
\cellcolor{TableColor2}&  \textbf{Duration} &  $\Delta t$ & $\,\left\|\vec{a}\right\|_{2}$ & Duration of movement bout. \\
\midrule 
\multirow[t]{2}{*}{\cellcolor{TableColor2}\textbf{Global}
                      } & \textbf{Clusters}   &  $\operatorname{Hopkins\,H}$, $\operatorname{KDE}_\textrm{moves}\rightarrow \operatorname{peaks+prom}$ & $\,\left\|\vec{a}\right\|_{2}$ & Degree of clustering or dispersion of movements throughout the night. \\
 \cellcolor{TableColor2}&  \textbf{Number of Movements}     &   $\operatorname{N}_\textrm{moves}/\textrm{h}$  & $\,\left\|\vec{a}\right\|_{2}$  & Overall activity during the night.\\
  \cellcolor{TableColor2} &  \textbf{Inter-Event Intervals}     &   $\operatorname{IEI}_\textrm{moves}$  & $\,\left\|\vec{a}\right\|_{2}$  & Movement spacing and irregularity across the night.\\
\bottomrule
\end{tabular}
\label{table:feature_list}
\end{table}

\newpage 

\section{Hyperparameter Search Space}\label{supp_c:hpsearchspace}
\markboth{Supplementary C}{}
\begin{table}[!ht]
\centering
\caption{\textbf{Hyperparameter Search Space for XGBoost}}
\setlength{\tabcolsep}{0.25cm}
\renewcommand{\arraystretch}{1.6}
\rowcolors{2}{TableColor2}{TableColor1}
\begin{tabular}{l l c r}
\toprule
\textbf{Name} & \textbf{Interpretation} & \textbf{Range} & \textbf{Type} \\
\midrule
n\_estimators      & Number of boosting trees                      & [300, 1100]         & Integer (uniform) \\
max\_depth         & Maximum individual tree depth                  & [6, 12]             & Integer (uniform) \\
min\_child\_weight & Minimum sum Hessian (min child weight)         & [1, 15]             & Real (log-uniform) \\
learning\_rate     & Shrinkage step size                            & [0.02, 0.12]        & Real (log-uniform) \\
subsample          & Row subsampling ratio                          & [0.55, 0.95]        & Real (uniform) \\
colsample\_bytree  & Feature subsampling per tree                   & 0.5                 & Fixed to default \\
colsample\_bylevel & Feature subsampling per tree level             & 0.3                 & Fixed to default  \\
reg\_alpha         & L1 regularization strength                     & 0.0                 & Fixed to default  \\
reg\_lambda        & L2 regularization strength                     & 1.0                 & Fixed to default  \\
top\_k\_feats      & Number of selected features (custom)           & [5, 35]             & Integer (uniform) \\
\bottomrule
\end{tabular}
\label{table:hps}
\end{table}

\section{Per Dataset Ranking Correlation}\label{supp_d:perdsranking}
\markboth{Supplementary D}{}
\begin{figure}[H] 
	\centering
	\includegraphics[page=1, width=\textwidth]{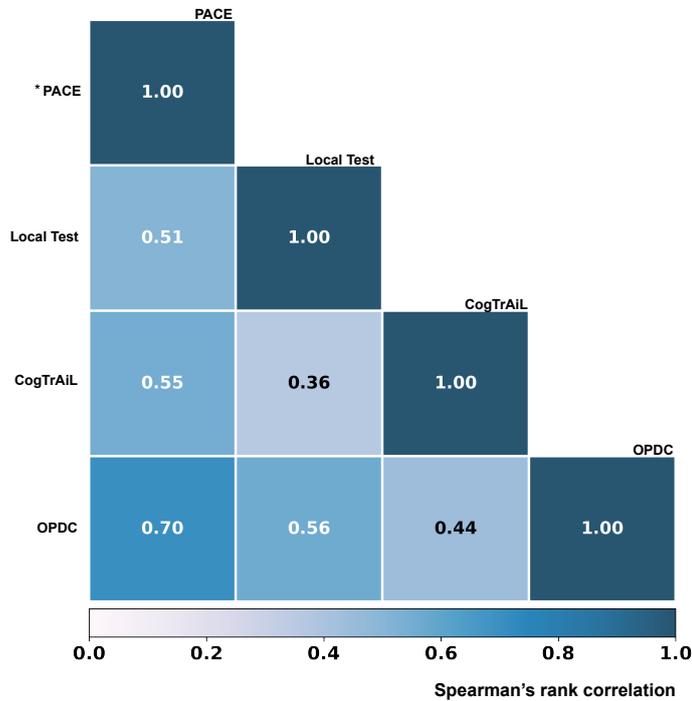} 
    \vspace{-.6cm}
	\caption{\label{fig:sup_per_dataset_rankings}
    \textbf{Spearman’s rank correlations of feature importance rankings derived within individual datasets.} Correlations ranged from 0.36 to 0.70, indicating moderate-to-strong agreement overall while still reflecting cohort-specific feature preferences. This variability underscores the value of pooling data to optimize the generalizability of feature sets. }
\end{figure}

\section{Multi-Center Model Validation (including \cgntest{})}\label{supp_e:multiwithcgntest}
\markboth{Supplementary E}{}
\begin{table}[htbp]
\centering
\caption{\textbf{Multi-Center Model Validation (including \cgntest{}).}
Displayed are classification metrics across LODO folds when \cgntest{} is part of the evaluation. 
The confidence intervals ($95\%$) are estimated via stratified bootstrap ($n=2000$). 
Upper block: all-RBD; lower block: iRBD-only.}
\setlength{\tabcolsep}{.1cm}
\renewcommand{\arraystretch}{1.6}
\rowcolors{2}{TableColor2}{TableColor1}
\begin{tabular}{>{\raggedleft}p{2.1cm}  >{\raggedleft}p{2.8cm} r c c c}
\toprule
\textbf{Cohort}  & \textbf{Records} & \textbf{AUROC} & $\mathbf{F_1}$ & \textbf{Bal. Acc.} \\
\toprule
\multirow[t]{3}{*}{\cellcolor{TableColor2}\textbf{\aar{}}} ({\textbf{\allrbd{}}}) 
& \mbox{}\hspace{2.1em}iRBD: \phantom{0}23 (171) \newline \pdrbd{}: \phantom{0}19 (117) \newline \mbox{}\hspace{.3em}\pdnorbd{}: \phantom{00}9 (\phantom{0}38) \newline HC: \phantom{00}6 (\phantom{0}57)
& \ciThree{0.8413}{0.83773}{0.84488}{0.08149} 
& \ciThree{0.88115}{0.87958}{0.88272}{0.03584} 
& \ciThree{0.70204}{0.69818}{0.70589}{0.08787} \\
\specialrule{0.04em}{0pt}{0pt}
\multirow[t]{2}{*}{\cellcolor{TableColor2}\textbf{\cgntest{}}} ({\textbf{\allrbd{}}}) 
& \mbox{}\hspace{2.1em}iRBD: \phantom{0}19 (119) \newline HC: \phantom{0}12 (\phantom{0}79)
& \ciThree{0.81306}{0.80955}{0.81658}{0.08012} 
& \ciThree{0.70304}{0.69892}{0.70716}{0.09401} 
& \ciThree{0.7214}{0.71789}{0.72491}{0.08002} \\
\specialrule{0.04em}{0pt}{0pt}
\multirow[t]{2}{*}{\cellcolor{TableColor2}\textbf{\cgntrainshort{}}} ({\textbf{\allrbd{}}}) 
& \mbox{}\hspace{2.1em}iRBD: \phantom{0}19 (119) \newline HC: \phantom{0}12 (\phantom{0}79)
& \ciThree{0.8952}{0.89354}{0.89685}{0.03777} 
& \ciThree{0.81168}{0.80979}{0.81356}{0.04289} 
& \ciThree{0.80973}{0.80781}{0.81164}{0.04366} \\
\specialrule{0.04em}{0pt}{0pt}
\multirow[t]{3}{*}{\cellcolor{TableColor2}\textbf{\oxfshort{}}} ({\textbf{\allrbd{}}}) 
& \mbox{}\hspace{2.1em}iRBD: \phantom{0}80 (496) \hspace{.4em}\pdrbd{}: \phantom{00}8 (\phantom{0}50) \newline HC: \phantom{0}25 (144)
& \ciThree{0.78168}{0.77875}{0.78462}{0.06686} 
& \ciThree{0.90299}{0.90221}{0.90376}{0.01765} 
& \ciThree{0.67949}{0.67705}{0.68193}{0.05559} \\
\midrule
\multirow[t]{2}{*}{\cellcolor{TableColor2}\textbf{\aar{}}} \mbox{}\hspace{2.1em} ({\textbf{iRBD}}) 
& \mbox{}\hspace{2.1em}iRBD: \phantom{0}23 (171) \newline HC: \phantom{00}6 (\phantom{0}57)
& \ciThree{0.97447}{0.97283}{0.97612}{0.0376} 
& \ciThree{0.95741}{0.95556}{0.95925}{0.04209} 
& \ciThree{0.96065}{0.95901}{0.96229}{0.03739} \\
\specialrule{0.04em}{0pt}{0pt}
\multirow[t]{2}{*}{\cellcolor{TableColor2}\textbf{\cgntest{}}} ({\textbf{iRBD}}) 
& \mbox{}\hspace{2.1em}iRBD: \phantom{0}19 (119) \newline HC: \phantom{0}12 (\phantom{0}79)
& \ciThree{0.74749}{0.74338}{0.7516}{0.09372} 
& \ciThree{0.65899}{0.65492}{0.66307}{0.09299} 
& \ciThree{0.63635}{0.63232}{0.64039}{0.09195} \\
\specialrule{0.04em}{0pt}{0pt}
\multirow[t]{2}{*}{\cellcolor{TableColor2}\textbf{\cgntrainshort{}}} ({\textbf{iRBD}}) 
& \mbox{}\hspace{2.1em}iRBD: \phantom{0}19 (119) \newline HC: \phantom{0}12 (\phantom{0}79)
& \ciThree{0.88211}{0.8803}{0.88391}{0.04121} 
& \ciThree{0.79033}{0.78836}{0.79231}{0.04496} 
& \ciThree{0.77772}{0.77559}{0.77985}{0.04859} \\
\specialrule{0.04em}{0pt}{0pt}
\multirow[t]{2}{*}{\cellcolor{TableColor2}\textbf{\oxfshort{}}} ({\textbf{iRBD}}) 
& \mbox{}\hspace{2.1em}iRBD: \phantom{0}80 (496) \newline HC: \phantom{0}25 (144)
& \ciThree{0.80598}{0.80319}{0.80876}{0.06345} 
& \ciThree{0.9077}{0.90699}{0.9084}{0.01613} 
& \ciThree{0.6914}{0.689}{0.6938}{0.05471} \\
\bottomrule
\end{tabular}
\footnotesize{}
\label{table:lodo_results_with_cgntest}
\end{table}

\newpage 
\section{Statistical Significances of Presented Features Per Dataset}\label{supp_f:statsignperds}
\markboth{Supplementary F}{}
\begin{table}[!ht]
\centering
\caption{\textbf{Statistical Significances of Presented Features Per Dataset.} P-values are computed using one-sided Mann-Whitney-U test and are presented alongside effect sizes, expressed as Cliff's Delta. The sample sizes (N) per dataset are given as RBD/HC. Statistically signifcant values (<0.05) are marked bold.}
\setlength{\tabcolsep}{0.1cm}
\renewcommand{\arraystretch}{1.6}
\begin{tabular}{>{\raggedleft}p{2.1cm} r  c  c  c  c  c}
\toprule
& & {\textbf{Pooled}} & {\textbf{\cgntrainshort{}}} & {\textbf{\cgntest{}}} & {\textbf{\oxfshort{}}} & {\textbf{\aar{}}} \\
 & & \footnotesize{N=1334/499} & \footnotesize{N=370/154}& \footnotesize{N=119/79} & \footnotesize{N=546/144} & \footnotesize{N=299/122} \\
\midrule
\multirow{2}{*}{\parbox{1.5cm}{\textbf{Power Skewness}}} & \footnotesize{p-value} & \cellcolor{TableColor1} $\mathbf{4.5\times 10^{-42}}$ & \cellcolor{TableColor1} $\mathbf{3.6\times 10^{-12}}$ & \cellcolor{TableColor1} $\mathbf{0.002}$ & \cellcolor{TableColor1} $\mathbf{3.1\times 10^{-19}}$ & \cellcolor{TableColor1} $\mathbf{2.2\times 10^{-9}}$ \\
& \footnotesize{Cliff's $\delta$} & \cellcolor{TableColor1} -0.41 & \cellcolor{TableColor1} -0.39 & \cellcolor{TableColor1} -0.26 & \cellcolor{TableColor1} -0.49 & \cellcolor{TableColor1} -0.37 \\
\midrule
\multirow{2}{*}{\parbox{1.5cm}{\textbf{First Autocorr. Minimum}}} & \footnotesize{p-value} & \cellcolor{TableColor2} $\mathbf{5.7\times 10^{-17}}$ & \cellcolor{TableColor2} $\mathbf{6.4\times 10^{-13}}$ & \cellcolor{TableColor2} $\mathbf{0.001}$ & \cellcolor{TableColor2} $\mathbf{3.1\times 10^{-8}}$ & \cellcolor{TableColor2} $\mathbf{7.8\times 10^{-10}}$ \\
& \footnotesize{Cliff's $\delta$} & \cellcolor{TableColor2} 0.25 & \cellcolor{TableColor2} 0.40 & \cellcolor{TableColor2} 0.28 & \cellcolor{TableColor2} 0.30 & \cellcolor{TableColor2} 0.38 \\
\midrule
\multirow{2}{*}{\parbox{1.5cm}{\textbf{Peak Frequency IQR}}} & \footnotesize{p-value} & \cellcolor{TableColor1} $\mathbf{2.9\times 10^{-7}}$ & \cellcolor{TableColor1} $\mathbf{1.0\times 10^{-12}}$ & \cellcolor{TableColor1} $\mathbf{4.2\times 10^{-5}}$ & \cellcolor{TableColor1} $0.660$ & \cellcolor{TableColor1} $\mathbf{0.019}$ \\
& \footnotesize{Cliff's $\delta$} & \cellcolor{TableColor1} -0.16 & \cellcolor{TableColor1} -0.40 & \cellcolor{TableColor1} -0.34 & \cellcolor{TableColor1} -0.02 & \cellcolor{TableColor1} -0.15 \\
\midrule
\multirow{2}{*}{\parbox{1.5cm}{\textbf{Median Bout Duration}}} & \footnotesize{p-value} & \cellcolor{TableColor2} $\mathbf{4.6\times 10^{-94}}$ & \cellcolor{TableColor2} $\mathbf{3.0\times 10^{-32}}$ & \cellcolor{TableColor2} $\mathbf{9.1\times 10^{-12}}$ & \cellcolor{TableColor2} $\mathbf{7.9\times 10^{-24}}$ & \cellcolor{TableColor2} $\mathbf{4.0\times 10^{-28}}$ \\
& \footnotesize{Cliff's $\delta$} & \cellcolor{TableColor2} 0.62 & \cellcolor{TableColor2} 0.66 & \cellcolor{TableColor2} 0.57 & \cellcolor{TableColor2} 0.54 & \cellcolor{TableColor2} 0.68 \\
\midrule
\multirow{2}{*}{\parbox{1.5cm}{\textbf{Spectral Entropy SD}}} & \footnotesize{p-value} & \cellcolor{TableColor1} $\mathbf{2.4\times 10^{-50}}$ & \cellcolor{TableColor1} $\mathbf{2.7\times 10^{-12}}$ & \cellcolor{TableColor1} $\mathbf{7.2\times 10^{-7}}$ & \cellcolor{TableColor1} $\mathbf{2.8\times 10^{-18}}$ & \cellcolor{TableColor1} $\mathbf{7.2\times 10^{-16}}$ \\
& \footnotesize{Cliff's $\delta$} & \cellcolor{TableColor1} -0.45 & \cellcolor{TableColor1} -0.39 & \cellcolor{TableColor1} -0.42 & \cellcolor{TableColor1} -0.47 & \cellcolor{TableColor1} -0.50 \\

\bottomrule
\end{tabular}
\label{table:feat_signifcances_per_dataset}
\end{table}

\newpage 

\section{Medication Prevalence Per Dataset}\label{supp_g:medication}
\markboth{Supplementary G}{}
\new{Medication metadata were collected across cohorts at the closest clinical encounter to actigraphy, not always contemporaneously with the actigraphy nights themselves, which is a limitation but reflects real-world recording conditions.\\
We grouped medication into potentially influential (\textbf{bold}) for (sleep) motor movements versus non-influential drugs ({non-bold}) based on prior literature demonstrating effects on REM sleep physiology, RSWA or (nocturnal) motor behavior in general\,\cite{standardsofpracticecommitteeBestPracticeGuide2010,mccarterTreatmentOutcomesREM2013, postumaAntidepressantsREMSleep2013,mccarterAntidepressantsIncreaseREM2015,louzadaZopicloneTreatInsomnia2021,zhouMovementDisordersAssociated2022,dolsPrevalenceManagementSide2013,kallweitPharmacologicalTreatmentsSleep2023,zhangAssociationMagnesiumIntake2022}:
\begin{itemize}
    \item{\textbf{Melatonin}: can alter REM physiology and is recommended for symptomatic RBD treatment.\,\cite{standardsofpracticecommitteeBestPracticeGuide2010, mccarterTreatmentOutcomesREM2013}}
    \item{\textbf{Clonazepam / Benzodiazepines}: Sedative/motor-suppressing; commonly used for RBD.\,\cite{standardsofpracticecommitteeBestPracticeGuide2010, mccarterTreatmentOutcomesREM2013}}
    \item{\textbf{Z-drugs}: Z-drugs alter sleep architecture (sleep stage timing, amount of REM, etc.) and have been associated with complex sleep-related behaviours, although no consistent direct effects on REM motor atonia have been demonstrated.\,\cite{louzadaZopicloneTreatInsomnia2021}  Examples: Zopiclone, Zolpidem, Doxylamine etc.}
    \item{\textbf{Antidepressants}: Associated with increased RSWA / RBD-like features in some individuals.\,\cite{postumaAntidepressantsREMSleep2013, mccarterAntidepressantsIncreaseREM2015} Examples: Citalopram, Sertraline, Mirtazapine, etc.}
    \item{\textbf{Mood Stabilizers / Antiseizure Meds}: Can have an influence on the central-nervous system (CNS); may cause tremor/sedation.\,\cite{zhouMovementDisordersAssociated2022,dolsPrevalenceManagementSide2013} Examples: Sodium Valproate, Lamotrigine, Lithium etc.}
    \item{\textbf{Other Sleep-  or CNS-active Meds}: Includes medications with central nervous system effects that can influence sleep architecture, arousal, or psychomotor function. Gabapentinoids such as pregabalin have been shown to alter sleep motor correlates and induce somnolence/ataxia, while CNS stimulants like methylphenidate and modafinil modulate dopaminergic/noradrenergic systems relevant to motor control and wakefulness.\,\cite{kallweitPharmacologicalTreatmentsSleep2023} Examples: Gabapentin, Pregabalin,  Methylphenidate, Modafinil, etc.}
    \item{Supplements / Alternative Medicine: General, mostly non-prescription supplememts or medication. Such are mostly studied for nutritional and sleep-quality effects, not directly on motor impact. For example, Magnesium supplementation has modest effects on sleep-quality and duration, but there's no strong evidence of direct motor system modulation at typical doses.\,\cite{zhangAssociationMagnesiumIntake2022} Examples: Vitamins, Magnesium, etc.}
    \item{Antiplatelet / Anticoagulants: Such medication primarily act peripherally or via metabolic pathways and do not affect the CNS. Hence, we assume no direct motor effects. Examples: Aspirin, Clopidogrel, Apixaban.}
    \item{Lipid-lowering: Assumed to a have neglible motor effects, demonstrated to have no infleunce on any cognitive domain.\,\cite{liEffectsCholesterolloweringDrugs2026} Examples: Atorvastatin, Ezetimibe, Rosuvastatin.}
    \item{Gastrointestinal: Act peripherally on the digestive system, no evidence on direct affect of central motor control or motor patterns during sleep, if at all, via only indirect via symptoms/arousals. Examples: Omeprazole, Lansoprazole, Gaviscon.}
    \item{Other System Meds: Other, non-specific meds like Pain / Inflammation, Urological, Antidiabetics etc. Similiar to Gastrointenstinal medication expected to have none, or negligble direct impact on nocturnal motor patterns. Examples: Ibuprofen, Paracetamol, Metformin, Levothyroxine, Tamsulosin etc.}
\end{itemize}
\noindent\begin{minipage}{\linewidth}
    {\footnotesize
    \centering
    \begin{threeparttable}
        \captionof{table}{\textbf{Medication prevalence across cohorts.} Values are patient-level counts per diagnosis subgroup. Medication is grouped into categories which are either considered as potentially motor-influential (\textbf{bold}) or not (non-bold).}
        \label{tab:medication}
        \setlength{\tabcolsep}{1.3pt}
        \renewcommand{\arraystretch}{1.7}
        \begin{tabular}{p{3cm} c c c c c c c c c c c}
            \cmidrule{2-12}
            \multirow{2}{*}{\textbf{}} & \multicolumn{2}{c}{\cellcolor{TableColor2}\textbf{\cgntrain{}}} & \multicolumn{2}{c}{\cellcolor{TableColor2}\textbf{\cgntest{}}} & \multicolumn{3}{c}{\cellcolor{TableColor2}\textbf{\oxfshort{}}} & \multicolumn{4}{c}{\cellcolor{TableColor2}\textbf{\aar{}}} \\
            \cmidrule{2-12}
             & \cellcolor{TableColor1}iRBD & \cellcolor{TableColor1}HC & \cellcolor{TableColor2}iRBD & \cellcolor{TableColor2}HC & \cellcolor{TableColor1}iRBD & \cellcolor{TableColor1}PD+RBD & \cellcolor{TableColor1}HC & \cellcolor{TableColor2}iRBD & \cellcolor{TableColor2}PD+RBD & \cellcolor{TableColor2}PD-RBD & \cellcolor{TableColor2}HC \\
            \midrule
            \textbf{Samples}\textsuperscript{a} & \cellcolor{TableColor1}55 & \cellcolor{TableColor1}{23} & \cellcolor{TableColor2}19 & \cellcolor{TableColor2}12 & \cellcolor{TableColor1}70 (80) & \cellcolor{TableColor1}8 (8) & \cellcolor{TableColor1}25 (25) & \cellcolor{TableColor2}13 (23) & \cellcolor{TableColor2}10 (19) & \cellcolor{TableColor2}5 (9) & \cellcolor{TableColor2}3 (6) \\
            \midrule
            \textbf{Melatonin} & \cellcolor{TableColor1}5 & \cellcolor{TableColor1}{\tiny{na}} & \cellcolor{TableColor2}{\tiny{-}} & \cellcolor{TableColor2}- & \cellcolor{TableColor1}21 & \cellcolor{TableColor1}2 & \cellcolor{TableColor1}{\tiny{-}} & \cellcolor{TableColor2}5 & \cellcolor{TableColor2}{\tiny{-}} & \cellcolor{TableColor2}{\tiny{-}} & \cellcolor{TableColor2}{\tiny{-}} \\
            \textbf{Clonazepam} / \newline\textbf{Benzodiazepines} & \cellcolor{TableColor1}1 & \cellcolor{TableColor1}{\tiny{na}} & \cellcolor{TableColor2}{\tiny{-}} & \cellcolor{TableColor2}{\tiny{-}} & \cellcolor{TableColor1}33 & \cellcolor{TableColor1}{\tiny{-}} & \cellcolor{TableColor1}{\tiny{-}} & \cellcolor{TableColor2}3 & \cellcolor{TableColor2}{\tiny{-}} & \cellcolor{TableColor2}{\tiny{-}} & \cellcolor{TableColor2}{\tiny{-}} \\
            \textbf{Z-drugs} & \cellcolor{TableColor1}{\tiny{-}} & \cellcolor{TableColor1}{\tiny{na}} & \cellcolor{TableColor2}{\tiny{-}} & \cellcolor{TableColor2}1 & \cellcolor{TableColor1}2 & \cellcolor{TableColor1}{\tiny{-}} & \cellcolor{TableColor1}{\tiny{-}} & \cellcolor{TableColor2}{\tiny{-}} & \cellcolor{TableColor2}1 & \cellcolor{TableColor2}{\tiny{-}} & \cellcolor{TableColor2}{\tiny{-}} \\
            \textbf{Antidepressants} & \cellcolor{TableColor1}3 & \cellcolor{TableColor1}{\tiny{na}} & \cellcolor{TableColor2}2 & \cellcolor{TableColor2}1 & \cellcolor{TableColor1}18 & \cellcolor{TableColor1}{\tiny{-}} & \cellcolor{TableColor1}3 & \cellcolor{TableColor2}{\tiny{-}} & \cellcolor{TableColor2}2 & \cellcolor{TableColor2}{\tiny{-}} & \cellcolor{TableColor2}{\tiny{-}} \\
            \textbf{Mood stabilizers} /\newline \textbf{Antiseizure Meds} & \cellcolor{TableColor1}2 & \cellcolor{TableColor1}{\tiny{na}} & \cellcolor{TableColor2}{\tiny{-}} & \cellcolor{TableColor2}2 & \cellcolor{TableColor1}1 & \cellcolor{TableColor1}{\tiny{-}} & \cellcolor{TableColor1}{\tiny{-}} & \cellcolor{TableColor2}{\tiny{-}} & \cellcolor{TableColor2}{\tiny{-}} & \cellcolor{TableColor2}{\tiny{-}} & \cellcolor{TableColor2}{\tiny{-}} \\
            \textbf{Other Sleep- or}\newline \textbf{CNS-active Meds} & \cellcolor{TableColor1}{\tiny{-}} & \cellcolor{TableColor1}{\tiny{na}} & \cellcolor{TableColor2}{\tiny{-}} & \cellcolor{TableColor2}{\tiny{-}} & \cellcolor{TableColor1}6 & \cellcolor{TableColor1}{\tiny{-}} & \cellcolor{TableColor1}1 & \cellcolor{TableColor2}1 & \cellcolor{TableColor2}1 & \cellcolor{TableColor2}1 & \cellcolor{TableColor2}{\tiny{-}} \\
            \specialrule{0.04em}{0pt}{0pt}
            $\mathbf{\sum\textrm{any}}$ \textbf{\textsuperscript{b}} & \cellcolor{TableColor1}10 & \cellcolor{TableColor1}{\tiny{na}} & \cellcolor{TableColor2}2 & \cellcolor{TableColor2}2 & \cellcolor{TableColor1}48 & \cellcolor{TableColor1}2 & \cellcolor{TableColor1}4 & \cellcolor{TableColor2}9 & \cellcolor{TableColor2}2 & \cellcolor{TableColor2}1 & \cellcolor{TableColor2}\tiny{-} \\
            \midrule
            Supplements / \newline Alternative Medicine & \cellcolor{TableColor1}14 & \cellcolor{TableColor1}{\tiny{na}} & \cellcolor{TableColor2}4 & \cellcolor{TableColor2}2 & \cellcolor{TableColor1}10 & \cellcolor{TableColor1}{\tiny{-}} & \cellcolor{TableColor1}2 & \cellcolor{TableColor2}\tiny{na} & \cellcolor{TableColor2}\tiny{na} & \cellcolor{TableColor2}\tiny{na} & \cellcolor{TableColor2}\tiny{na} \\
            Antiplatelet / \newline Anticoagulants & \cellcolor{TableColor1}12 & \cellcolor{TableColor1}{\tiny{na}} & \cellcolor{TableColor2}5 & \cellcolor{TableColor2}1 & \cellcolor{TableColor1}17 & \cellcolor{TableColor1}3 & \cellcolor{TableColor1}3 & \cellcolor{TableColor2}\tiny{na} & \cellcolor{TableColor2}\tiny{na} & \cellcolor{TableColor2}\tiny{na} & \cellcolor{TableColor2}\tiny{na} \\
            Lipid-lowering & \cellcolor{TableColor1}17 & \cellcolor{TableColor1}{\tiny{na}} & \cellcolor{TableColor2}4 & \cellcolor{TableColor2}3 & \cellcolor{TableColor1}28 & \cellcolor{TableColor1}1 & \cellcolor{TableColor1}13 & \cellcolor{TableColor2}\tiny{na} & \cellcolor{TableColor2}\tiny{na} & \cellcolor{TableColor2}\tiny{na} & \cellcolor{TableColor2}\tiny{na} \\
            Gastrointestinal & \cellcolor{TableColor1}8 & \cellcolor{TableColor1}{\tiny{na}} & \cellcolor{TableColor2}1 & \cellcolor{TableColor2}1 & \cellcolor{TableColor1}20 & \cellcolor{TableColor1}1 & \cellcolor{TableColor1}9 & \cellcolor{TableColor2}\tiny{na} & \cellcolor{TableColor2}\tiny{na} & \cellcolor{TableColor2}\tiny{na} & \cellcolor{TableColor2}\tiny{na} \\
            Other Systemic Meds & \cellcolor{TableColor1}38 & \cellcolor{TableColor1}{\tiny{na}} & \cellcolor{TableColor2}12 & \cellcolor{TableColor2}7 & \cellcolor{TableColor1}\tiny{na} & \cellcolor{TableColor1}\tiny{na} & \cellcolor{TableColor1}\tiny{na} & \cellcolor{TableColor2}\tiny{na} & \cellcolor{TableColor2}\tiny{na} & \cellcolor{TableColor2}\tiny{na} & \cellcolor{TableColor2}\tiny{na} \\
            \bottomrule
        \end{tabular}
        \vspace{0.5em}
        \begin{tablenotes}[flushleft]
            \footnotesize
            \item \textbf{na}: Data not available.
            \item $^\mathbf{a}$ \textbf{Number of contributed recordings}: For \cgntrain{} and \cgntest{}, this is equal to the number of unique subjects and actigraphy recordings. For \aar{} and \oxfshort{}, its equal to $\textrm{N}_\textrm{subjects}\left(\textrm{N}_\textrm{records}\right)$, as some subjects contributed multiple recordings. For \aar{}, only recordings with $\geq 4$ nights are shown.
             \item $^\mathbf{b}$ \textbf{Number of subjects with any motor-influential medication}: Count of subjects with medication from any of the as potentially motor-influential classified categories, some subjects have medication in multiple categories. 
        \end{tablenotes}
    \end{threeparttable}}
\end{minipage}
\vspace{.1cm}\newline
Across cohorts, potentially motor-influential medication had higher prevalence within RBD-positive subgroups.
When considering \emph{any} motor-influential medication, prevalence varied across cohorts. In \oxfshort{}, 50 of 70 RBD positive subjects (iRBD \& \pdrbd{}) and 2 of 25 HC subjects had at least one motor-influential medication recorded. In \aar{}, 11 of 23 RBD positive and 1 of 8 RBD negative (HC \& \pdnorbd{}) had at least one such medication. In \cgntest{}, motor-influential medication was observed in both RBD (2/19) and HC (2/12) participants. \cgntrain{} only recorded motor-influential medication for iRBD subjects (10/55), with no HC medication data available.\\
}

\end{appendices}

\newpage
\markboth{References}{}
\bibliography{references}

\end{document}